\documentclass[10pt, letter, onecolumn]{arxiv}

\usepackage{kantlipsum, lipsum}
\usepackage{dm-colors}
\usepackage{amsmath}
\usepackage{pstricks, pst-node}
\usepackage{verbatim}
\usepackage{multirow}
\usepackage{scalerel}
\usepackage{booktabs}
\usepackage{enumitem}
\usepackage{xspace}
\usepackage{bm}
\usepackage{bbm}
\usepackage{mathtools}
\usepackage{soul}
\usepackage{epsfig}
\usepackage{graphicx}
\usepackage{tcolorbox}
\usepackage{subcaption}
\usepackage{amssymb}
\usepackage{colortbl}
\usepackage{csquotes}
\usepackage{etex}
\usepackage{setspace}
\usepackage{svg}
\usepackage{colortbl}
\usepackage{tabularx,ragged2e}
\usepackage{placeins}
\usepackage[symbol]{footmisc}
\usepackage[bibstyle=nature,citestyle=numeric-comp,%
            natbib=true,backend=biber,maxbibnames=99,%
            giveninits=false,sorting=none]{biblatex}
\usepackage{nameref}
\usepackage{varioref}
\usepackage[pagebackref=false,breaklinks=false,%
            colorlinks=true,bookmarks=true,citecolor=ourdarkblue,%
            urlcolor=ourdarkblue,linkcolor=ourdarkblue]{hyperref}
\usepackage[noabbrev,capitalize]{cleveref}
\usepackage{etoc}

\graphicspath{{figures/}}

\def\eg{{\em e.g.,~}}
\def\ie{{\em i.e.,~}}

\def\etal{{\em et al.}\xspace}
\def\ourmodel{{Tx-LLM}\xspace}
\def\basemodel{{PaLM 2}\xspace}
\def\ourmodels{{\ourmodel (S)}\xspace}
\def\ourmodelm{{\ourmodel (M)}\xspace}
\def\basemodels{{\basemodel (S)}\xspace}
\def\basemodelm{{\basemodel (M)}\xspace}
\def\ourdata{{TxT}\xspace}
\def\ndatasets{{709}\xspace}
\def\ntasks{{66}\xspace}
\def\nbetterthansota{{22}\xspace}
\def\nbinarybetterthansota{{12}\xspace}
\def\nregressionbetterthansota{{10}\xspace}
\def\natleastnearsota{{43}\xspace}

\def\nregressionatleastnearsota{{19}\xspace}
\def\nnearsota{{21}\xspace}
\def\nbinarynearsota{{12}\xspace}
\def\nregressionnearsota{{9}\xspace}

\title{Tx-LLM: A Large Language Model for Therapeutics}

\author[$\ast$,1]{Juan Manuel Zambrano Chaves}
\author[$\ast$,2]{Eric Wang}
\author[2]{Tao Tu}
\author[$~$\hspace{-4pt}]{Eeshit Dhaval Vaishnav}
\author[$~$\hspace{-3pt}]{Byron Lee}
\author[2]{S. Sara Mahdavi}
\author[1]{Christopher Semturs}
\author[2]{David Fleet}
\author[$\dagger$,1]{\\Vivek Natarajan}
\author[$\dagger$,$\ddagger$,2]{Shekoofeh Azizi}

\affil[1]{Google Research, }
\affil[2]{Google DeepMind }

\renewcommand{\correspondingauthor}[1]{$\ast$~Equal contributions. Contributions made as a student researcher. %
                                       $\dagger$~Equal leadership. \\%
                                       $\ddagger$~Corresponding author: shekazizi@google.com }

\begin{document}
\begin{refsection}

\begin{abstract}
Developing therapeutics is a lengthy and expensive process that requires the satisfaction of many different criteria, and AI models capable of expediting the process would be invaluable. However, the majority of current AI approaches address only a narrowly defined set of tasks, often circumscribed within a particular domain. To bridge this gap, we introduce \ourmodel, a generalist large language model (LLM) fine-tuned from PaLM-2 which encodes knowledge about diverse therapeutic modalities. \ourmodel is trained using a collection of \ndatasets datasets that target \ntasks tasks spanning various stages of the drug discovery pipeline. Using a single set of weights, \ourmodel simultaneously processes a wide variety of chemical or biological entities (small molecules, proteins, nucleic acids, cell lines, diseases) interleaved with free-text, allowing it to predict a broad range of associated properties, achieving competitive with state-of-the-art (SOTA) performance on \natleastnearsota out of \ntasks tasks and exceeding SOTA on \nbetterthansota. Among these, \ourmodel is particularly powerful and exceeds best-in-class performance on average for tasks combining molecular SMILES representations with text such as cell line names or disease names, likely due to context learned during pretraining. We observe evidence of positive transfer between tasks with diverse drug types (\eg tasks involving small molecules and tasks involving proteins), and we study the impact of model size, domain finetuning, and prompting strategies on performance. We believe \ourmodel represents an important step towards LLMs encoding biochemical knowledge and could have a future role as an end-to-end tool across the drug discovery development pipeline.
\end{abstract}

\maketitle


\section{Introduction}
\label{sec:introduction}

Developing therapeutics is a risky enterprise, as 90\% of clinical trial candidates fail, and even successful therapeutics typically take 10-15 years and 1-2 billion dollars until approval~\cite{sun_why_2022,hinkson_accelerating_2020}. Perhaps the most daunting obstacle in this process is that a successful therapeutic must simultaneously satisfy numerous criteria. For example, a drug should interact with its proposed target, ultimately leading to the desired therapeutic effect and clinical efficacy. At the same time, the drug should be non-toxic and have drug-like properties (\eg solubility, permeability, suitable pharmacokinetics, and pharmacodynamics). In clinical trials, unexpected off-target effects and interactions may counterbalance the effects of an otherwise promising drug candidate~\cite{lin_off-target_2019}. Furthermore, practical considerations regarding small molecule synthesis, or biological molecule developability must be taken into account.

\begin{figure}[]
    \centering
    \includegraphics[width=\textwidth]{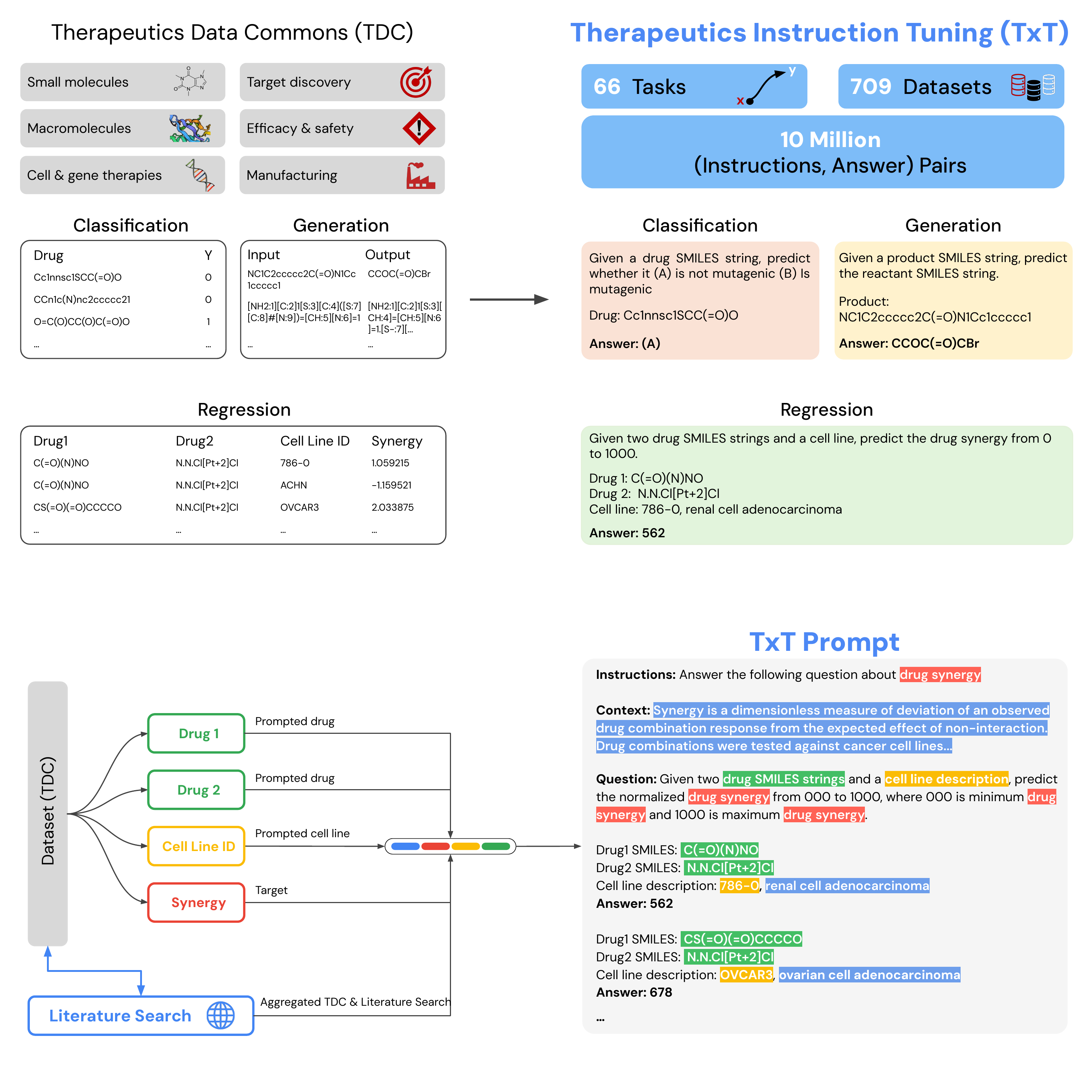}
    \caption{\textbf{Overview of the \ourmodel.} \textbf{(top)} Datasets from the Therapeutic Data Commons are used to construct the Therapeutics instruction Tuning (\ourdata) collection. The original tabular datasets contain a variety of drug types including small molecules, macro-molecules such as proteins and nucleic acids, cells, and genes. The tasks encompass a broad range of areas relevant to drug discovery and development such as predicting targets, evaluating efficacy and safety, and predicting ease of manufacturing. \ourdata interleaves free-text instructions with string representations of molecules, such as SMILES strings for small molecules or amino acid sequences for proteins. \ourdata is used to prompt and finetune \ourmodel to solve classification, regression, or generation tasks.
    \color{black}\textbf{(bottom)} Example of a \ourdata prompt for predicting drug synergy. The prompt is composed of Instructions, Context, and a Question using information from the corresponding TDC dataset and/or literature search and may also contain exemplars to aid in-context learning.}
    \label{fig:overview}
\end{figure}

Given the expense of experimentally assessing each of these characteristics, curated collections of experimental data paired with machine learning models for predicting therapeutic properties are useful as initial screening steps~\cite{sadybekov_computational_2023}. To aid the development of these models, the Therapeutics Data Commons (TDC) was developed as a resource containing AI-ready datasets and benchmarks for a diverse range of therapeutics-related tasks such as drug-target binding prediction or drug toxicity prediction~\cite{huang_artificial_2022,huang2021therapeutics}. Current state-of-the-art (SOTA) models for TDC datasets are largely focused on individual tasks with one approach being to train a library of specialist models and call upon different specialists for each step of the therapeutic development pipeline~\cite{huang_artificial_2022}. However, specialist models lack awareness of other tasks in the therapeutic development pipeline, which may in turn limit their ability to contextualize and improve performance.

Large language models (LLMs) have emerged as useful systems for encoding information from large-scale data and communicating the information using language. Interestingly, LLMs have shown promise for multiple types of tasks such as multiple-choice question answering \cite{robinson2023leveraging}, time series prediction \cite{gruver2023large}, and regression \cite{song2024omnipred}. Furthermore, LLMs can be effectively adapted to specific domains such as medicine or chemistry with in-context learning or finetuning \cite{singhal_large_2023,singhal2023expertlevel,tu2023generalist,zhang_fine-tuning_2024,jablonka_leveraging_2024}. We hypothesized that tasks across the therapeutic development pipeline, even those involving diverse drug types such as small molecules and protein sequences, could be combined to train a generalist LLM with improved performance on individual tasks while using the same set of weights for all tasks.

\begin{figure}
    \centering
    \includegraphics[width=0.8\textwidth]{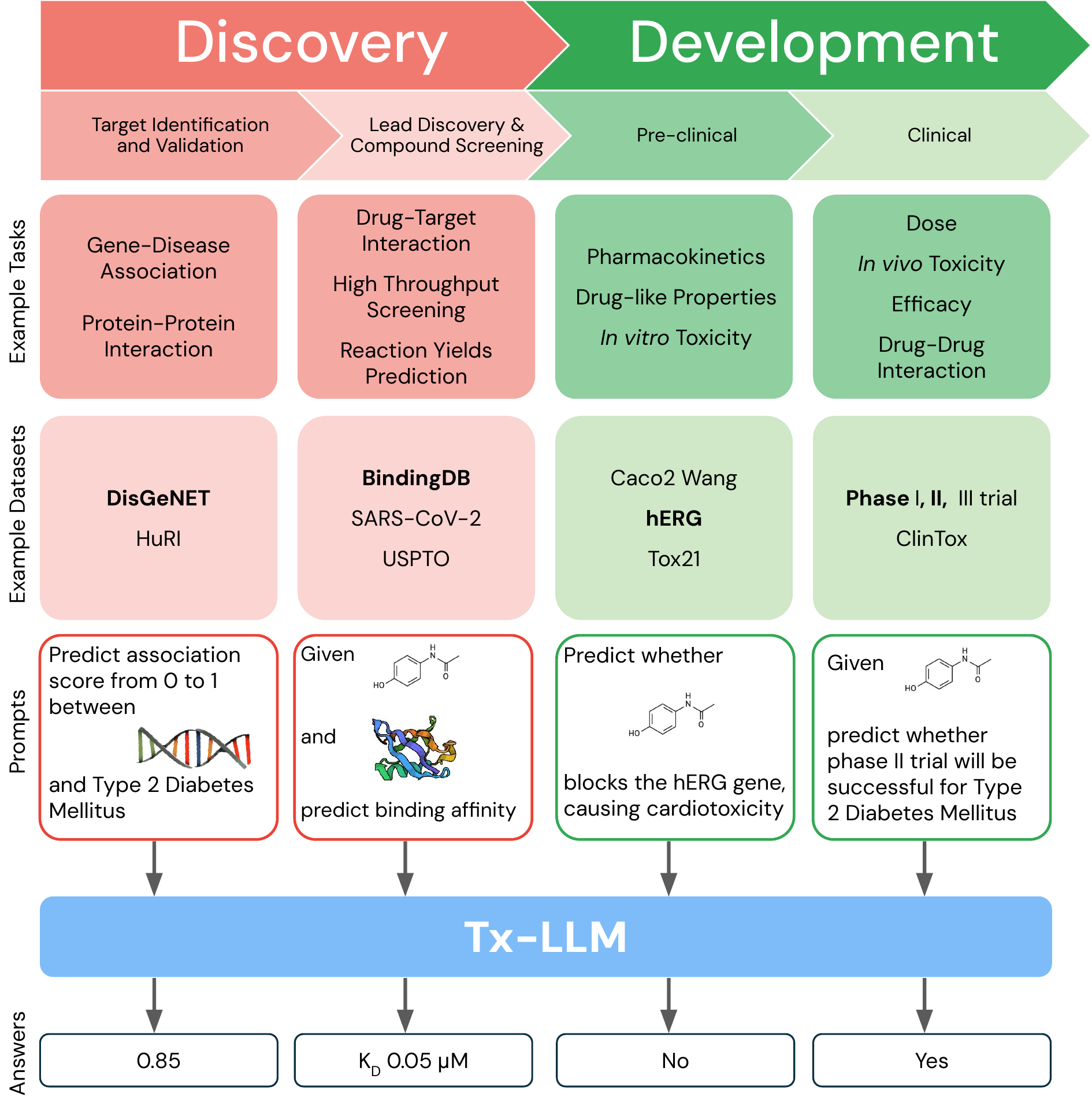}
    \vspace*{6pt}
    \caption{\textbf{\ourmodel may \textcolor{black}{be effective} for end-to-end therapeutic development.} \ourmodel is a single model that can be queried for multiple steps of the therapeutic development process, covering tasks from early-stage target discovery to late-stage clinical trial approval. We list example tasks associated with each stage of the therapeutic development pipeline, example datasets in TDC that correspond to these tasks, and example prompts that can be used to query \ourmodel. For illustration, the example prompts are geared towards discovering new small molecules against targets associated with type 2 diabetes, and the datasets associated with the example prompts are shown in bold.}
    \label{fig:potential_use}
\end{figure}

In this work, we develop and introduce such a generalist model, \ourmodel, by representing therapeutics as strings and finetuning the PaLM-2 base LLM on a diverse set of classification, regression, and generation tasks (\cref{fig:overview}). Our key contributions are as follows:

\begin{itemize}
    \item \ourmodel performs above or near SOTA for \natleastnearsota out of \ntasks tasks. For datasets combining molecular string representations with text, \ourmodel is especially effective and exceeds SOTA on average for these, likely due to context learned during LLM pretraining.
    \item Interestingly, we find evidence of positive transfer between datasets with diverse drug types, as training on datasets including biological sequences improves performances on molecular datasets.
    \item We perform ablation studies and observe that scale, domain finetuning, and prompting strategies also significantly impact model performance.
    \item \textcolor{black}{The proposed \ourmodel shows promise as an end-to-end therapeutic development assist, allowing one to query a single model for multiple steps of the development pipeline \textcolor{black}{(\cref{fig:potential_use})}}.
\end{itemize}

\section{Related works}

\textbf{Large language models (LLMs)} Since the advent of transformer-based models \cite{vaswani2023attention}, LLMs have become increasingly powerful at a variety of natural language processing tasks \cite{NEURIPS2020_1457c0d6,JMLR:v21:20-074}. LLMs are trained using self-supervised learning on large-scale text corpi and have been shown to encode information while also generalizing to unseen tasks. Interestingly, it has recently been shown that LLMs are able to perform regression on diverse tasks using only textual representations of mathematical parameters and values \cite{song2024omnipred}.

\textbf{Specialist models for therapeutics} Therapeutics have been represented in a variety of ways. Molecules can be naturally represented as graphs, and graph neural networks (GNNs) have been applied for a variety of prediction or generation tasks \cite{torng_graph_2019,stark2022equibind,xiong_pushing_2020,heid_machine_2022,yang_analyzing_2019,morrone_docking,mohr_data-driven_2022}. A notable application of GNNs was the discovery of Halicin, an antibiotic which was effective against previously pan-resistant bacterial strains \cite{stokes_deep_2020}. Molecules are also commonly represented using binary vectors (fingerprints), which capture the local environment of each atom in the molecule within a predefined radius and can be input into a variety of models \cite{ran_drug-drug_2022,tayyebi_prediction_2023,belenahalli_shekarappa_development_2023}. Proteins and nucleic acids are conveniently represented using their amino acid or nucleotide sequences, and which can then be encoded in multiple ways to predict properties such as fitness \cite{louie_gp160}, binding \cite{MHC1_IEDB_IMGT_Nielsen_ref, MHC2_IEDB_Jensen_ref}, and secondary structure \cite{wang_attfold:_2020}. The introduction of AlphaFold and its successors significantly advanced the field of structure prediction \cite{jumper_highly_2021,tunyasuvunakool_highly_2021,senior_improved_2020,abramson_accurate_2024}, which may allow further developments in areas such as design of structure-based drugs \cite{ren_alphafold_2023} and vaccines \cite{higgins_can_2021}.

\textbf{Language models for biology and chemistry} Language models have also emerged as tools for biology and chemistry. While many LLMs have shown limited performance for chemistry \cite{white_digital_discovery,guo2023what}, LlaSMol recently showed that finetuning LLMs for chemistry achieved near-SOTA performance on multiple tasks \cite{yu2024llasmol}. Beyond chemical property prediction, Chemcrow used LLMs for robotics-guided organic synthesis without human interaction \cite{m._bran_augmenting_2024}. Protein language models such as ESM \cite{rives2019biological,lin_evolutionary-scale_2023}, Unirep \cite{alley_unified_2019}, and ProtGPT2 \cite{ferruz_protgpt2_2022} have used self-supervised pretraining (such as masked token prediction) to generate protein embeddings that were useful for downstream tasks such as predicting stability and functional effects of mutations. It has also been proposed that the probabilities assigned to amino acids during masked token prediction correlate to fitness \cite{doi:10.1126/science.abd7331} and provide an efficient landscape for computational protein engineering \cite{hie_efficient_2024}. ProtLLM combined protein sequence encoders with language encoders \cite{zhuo2024protllm}, and BioT5 combined molecular string representations with protein names, sequences, or structures to train LLMs on various prediction tasks \cite{pei2024biot5}. At the cellular level, scGPT \cite{cui_scgpt:_2024}, GenePT \cite{Chen2023.10.16.562533}, and Geneformer \cite{theodoris_transfer_2023} represent cells as rank ordered lists of highly expressed genes for tasks such as cell type annotation or therapeutic target discovery.
\section{Methods}
\label{sec:methods}

\subsection{Datasets}
\label{subsec:datasets}

We assembled \ourdata, a collection of \ndatasets drug discovery datasets comprising \ntasks tasks formatted for instruction tuning. We sourced our datasets from TDC, a publicly available repository that offers a wide variety of tasks spanning the drug discovery process. The median dataset size was 11,000, and the distribution of dataset sizes is illustrated in \cref{fig:dataset_size}. We excluded a small number of datasets found in TDC for various reasons, which are detailed in \cref{tab:excluded_results}.

Each dataset in \ourdata is formatted as a text prompt comprised of four components (instructions, context, question, answer), illustrated in \cref{fig:overview,tab:example_prompts_binary,tab:example_prompts_regression_generation,tab:example_prompts_fewshot}. Instructions consisted of a short sentence describing the task at hand, such as \textit{``Answer the following question about drug properties''}. For each dataset, we crafted context, \ie free-text descriptions providing additional information that grounds the question in a relevant biochemical setting. Contexts were 2-3 sentences long, sourced from TDC dataset descriptions, and manually complemented based on a brief literature search of the topic. For specialized assays describing a specific experimental condition, such as ToxCast, additional information for contexts were obtained from publicly available assay descriptions \cite{pubchem, toxcast_summary}. The question is a succinct query that specifies the specific property being asked, and interleaves English text with text-based representations of therapeutics (\eg \textit{``Does the following molecule cross the blood brain barrier? <molecule>''}. The format of answers varied depending of the type of task.

Datasets in \ourdata fell into one of three categories:
\begin{enumerate}[label=(\roman*)]
    \item \textbf{Binary classification} questions were formatted as a prediction of a single property of a therapeutic with two possibilities, yes/no (\eg whether a drug is toxic).
    \item \textbf{Regression} questions were formatted as a prediction of a single property of a therapeutic on a continuous scale (e.g. drug-target binding affinity). To leverage the token-based, and not float-based, representation in existing language models, we uniformly binned the labels between 0 and 1000 and instruct \ourmodel to predict the bin label. On evaluation, the predicted bin was transformed back to the original numeric label space.
    \item \textbf{Generation} we focused on one generation task, which consists of predicting reactants of a chemical reaction given the product, sourced from the USPTO dataset \cite{Lowe2017}. 
\end{enumerate}

String representations of diverse types of therapeutics in \ourdata fell into one of the following categories:
\begin{enumerate}[label=(\roman*)]
    \item \textbf{SMILES} Small molecules were represented with their SMILES string.
    \item \textbf{Amino acid} Proteins and peptides were represented with their amino acid sequences. Multiple Histopatibility Complex molecules, such as those found in the in the MHC1 IEDB IMGT Nielsen \cite{nielsen_netmhcpan-3.0;_2016} and MHC2 IEDB Jensen \cite{jensen_improved_2018} datasets were represented using their pseudo-sequences (only showing residues that are in contact with a peptide), T cell receptors were represented using their CDR3 hypervariable loops. 
    \item \textbf{Nucleotide} Nucleic acids were represented with their nucleotide sequence.
    \item \textbf{Amino acid + SMILES} Multi-instance datasets containing both proteins and small molecules used the protein amino acid sequence and the molecular SMILES string.
    \item \textbf{Nucleotide + Amino acid} Multi-instance datasets containing both nucleic acids and proteins used the nucleotide sequence and protein amino acid sequence.
    \item \textbf{SMILES + Text} Multi-instance datasets containing small molecules and other feature types used the molecular SMILES string and English text to represent the additional features, such as disease or cell line names and descriptions. Notable datasets in this category include Phase I, Phase II, and Phase III clinical trial datasets. These datasets contain information about the SMILES strings of candidate drugs, the names of targeted diseases for various clinical trial phases, and whether the trial ultimately received approval.
    \item \textbf{Amino acid + Text} Multi-instance datasets containing proteins and other feature types used the protein amino acid sequence and text to represent the other features. DisGeNET, which contained protein sequences and disease names, was the only dataset in this category.
\end{enumerate}

For each dataset, data splits were constructed using TDC functions with recommended split methods (random, scaffold, cold-start, combination, temporal), which are indicated in \cref{tab:binary_results,tab:regression_generation_results}. For datasets in the ADMET, DrugCombo, or DTI DG leaderboards, we followed the leaderboard-specific instructions for generating splits with a seed of 1.

\subsection{Modeling}

\textbf{Base LLM} \ourmodel was initiated from PaLM-2 \cite{anil2023palm}, the second generation of Google's LLM trained using the Pathways accelerator orchestration system \cite{barham2022pathways}. PaLM-2 models used in this work were trained at sizes S and M.

\textbf{Few-shot prompting} Few-shot prompting \cite{NEURIPS2020_1457c0d6} involves providing example inputs and outputs in the prompt. Based on prior evidence showing the benefits of using a mixture of 0 and few-shot tasks \cite{longpre2023flan}, we constructed \ourdata as a mixture of of 70\% 0-shot and 30\% few-shot prompts with the number of few shots randomly chosen between 1 and 10. If the shots caused the prompt to exceed the maximum length, then the number of shots was reduced until the length was below the maximum. The shots were selected using random datapoints in the training dataset. On evaluation, we also considered nearest neighbor shots, but we strictly used random shots during training because the datapoints were often more similar within the training set than across training and test sets (\cref{fig:nn_distribution}).

\textbf{Finetuning} We finetuned a non-instruction tuned variant of PaLM-2 using \ourdata training data. We trained a single model across all TDC datasets using dataset mixture ratios proportional to the number of datapoints in each dataset. \ourmodel generally refers to \ourmodelm models trained across all TDC datasets. We explored various key ablations using subsets of TDC datasets for comparison using a smaller, \ourmodels due to constraints on computational resources. Hyperparameters for finetuning are listed in \cref{tab:hyperparameters}.

\subsection{Evaluation}

\textbf{State-of-the-art performance} SOTA values reported in this work are primarily derived from academic literature. For TDC datasets included in the ADMET, DrugCombo, and DTI DG leaderboards, SOTA values were obtained directly from the respective leaderboard. For datasets not featured in a leaderboard, SOTA values were determined through a manual literature review.

\textbf{Few-shot prompting} We evaluated the performance of \ourmodel with 0 and few-shot prompting, varying the number of shots and whether the shots were selected based on random datapoints or nearest neighbor datapoints. Nearest neighbors were identified using representations of the query therapeutic(s). For small molecules, nearest neighbors were identified using Tanimoto similarities with Morgan fingerprints, whereas for amino acid and nucleotide sequences percent sequence identities were used. Morgan fingerprints and Tanimoto similarities were calculated using RDKit and Chemfp \cite{Landrum2016RDKit2016_09_4,dalke_chemfp_2019}, and multiple sequence alignments for percent sequence identities were calculated using Clustal Omega \cite{sievers_fast_2011}. For validation, shots were selected from the training sets, whereas shots for test sets were selected from the combined training and validation sets.

\textbf{Metrics} We report \ourmodel performances on TDC datasets using the preferred metric for each task as defined by \cite{huang2021therapeutics}. Metrics for binary classification datasets include area under the receiver operating characteristic curve (AUROC), area under the precision-recall curve (AUPRC), and accuracy. Metrics for regression datasets include the Spearman correlation coefficient, Pearson correlation coefficient, mean absolute error (MAE), and mean squared error (MSE). The metric for the USPTO generation dataset is set accuracy, where for each reaction set overlaps between the generated reactants compared to the ground-truth reactants were assigned a score of 1 if there was perfect overlap, and 0 otherwise.

\textbf{Statistical tests} Given performances on TDC datasets from two models, we sought to determine whether one model was significantly better considering all datasets. To do this, we performed a non-parametric Wilcoxon signed-rank test on the performances from both models and report the p-value. In order to account for the differences in magnitudes for MAE and MSE metrics, we normalized all performances by the mean of the performances from both models. We also reversed the sign of MAEs and MSEs because lower MAEs and MSEs correspond to better performances. 

\textbf{Data contamination analysis} \textcolor{black}{PaLM-2 was trained on diverse data~\cite{anil2023palm} that may contain information about therapeutics in TDC. To study this, we analyzed the percent overlap between TDC dataset features and the PaLM-2 training data. To illustrate an example, consider the case of a TDC dataset containing SMILES strings for molecules paired with target amino acid sequences. For each TDC datapoint, two searches over the PaLM-2 training data were performed using (i) the full SMILES string and (ii) the full amino acid sequence, and the datapoint was considered overlapping if either the SMILES string or amino acid sequence were found in the PaLM-2 training data. The minimum match length was set to the SMILES/amino acid sequence length, up to a maximum of 512 characters due to infrastructural constraints. Our analysis looks for direct character overlaps and does not account for different representations of the same feature, such as molecules being represented as SMILES strings in TDC but being referred to by name in the PaLM-2 training data.}
\section{Results}
\label{sec:results}

\subsection{Performance on TDC datasets}

\begin{figure}
    \centering
    \includegraphics[width=0.9\textwidth]{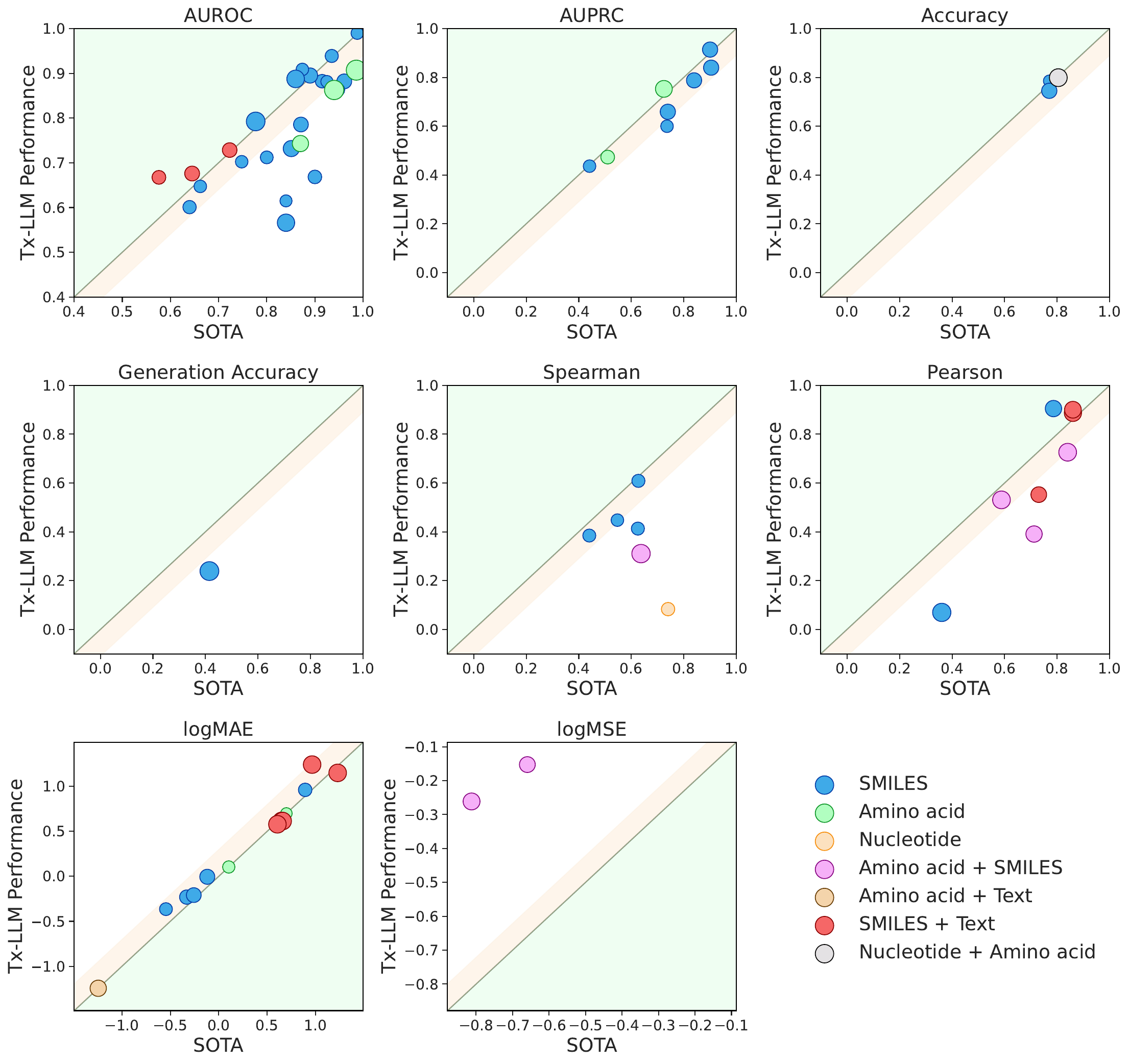}
    \vspace{6pt}
    \caption{\textbf{Comparison of \ourmodel's performance with SOTA.} \ourmodel is evaluated on each dataset in TDC, and comparison with SOTA for different metrics is illustrated in panels. Datasets are colored by their feature types indicated in the legend, and marker sizes illustrate the number of data points in the task on a log scale. The larger shaded area in green indicates where \ourmodel outperforms SOTA, while the narrower orange shaded area indicates where \ourmodel is near SOTA (defined as within 10\%). MAE and MSE values are log-transformed because the magnitudes of these values depend on the units of the outputs. Generation accuracy is the fraction of correct SMILES strings in the USPTO generation task.}
    \label{fig:tdc_results}
\end{figure}

The performance of \ourmodel on TDC datasets is summarized in \cref{fig:tdc_results,tab:binary_results,tab:regression_generation_results}. Out of \ntasks TDC tasks, \ourmodel performed near or exceeding SOTA on \natleastnearsota datasets. From these \natleastnearsota datasets, \ourmodel outperformed SOTA on \nbetterthansota (\nbinarybetterthansota binary classification datasets and \nregressionbetterthansota regression datasets) and performed near SOTA (defined as within 10\% of SOTA) for another \nnearsota (\nbinarynearsota binary classification datasets and \nregressionnearsota regression datasets). Notably, these results were achieved using the same set of model weights without any task-specific optimizations.

\textbf{\ourmodel is particularly effective at combining SMILES and text} TDC datasets containing features involving SMILES strings for molecules and text for other features (such as disease name or cell line name and description) tended to perform near or exceeding SOTA more frequently than datasets containing other features types (\cref{fig:tdc_results,tab:binary_results,tab:regression_generation_results}). To quantify this trend, we calculated the median relative difference of \ourmodel performance from SOTA (defined as (\ourmodel performance - SOTA) / SOTA) for each feature type (SMILES + Text, Nucleotide + Amino acid, SMILES, Amino acid, Amino acid + SMILES, Nucleotide) in \cref{tab:task_type_results}. In calculating the relative differences, signs were reversed for MAE and MSE metrics because lower MAEs and MSEs correspond to better performances. Amino acid + Text was not included because only the DisGeNET dataset, which we did not find a SOTA for, had this feature type. 

SMILES + Text was the only feature type yielding a positive median relative difference, suggesting that it was the only feature type which tended to exceed SOTA on average. This performance may be due to the text representations for diseases and cell lines, both because text is a natural representation for a LLM and because the base LLM may have learned context about these in its pretraining. The ability to exceed SOTA may also be in part due to the difficulty in representing these features with non-LLM models. For example, SOTA models for the clinical trial approval datasets (phase 1, phase 2, and phase 3) represent diseases as nodes in an interaction graph \cite{phase1_ref}, which may contain less information than context learned from LLM pretraining. 

Overall, these results suggest that finetuned LLMs may be particularly effective for tasks involving both a drug and a target that can be represented in text (such as disease name or cell line name). In a sense, a finetuned LLM is an intermediate model between a domain-specific model, such as a GNN which is effective for representing molecules but less effective for other features, and a base LLM that does not understand molecular SMILES strings \cite{white_digital_discovery} but does contain diverse knowledge about physiology. Thus, a finetuned LLM may be an ideal model for tasks involving both.

\textbf{Limitations of \ourmodel for datasets without textual features} In contrast, \ourmodel underperformed SOTA on small molecule datasets solely using SMILES strings (\cref{tab:task_type_results}). This suggests that SOTA models representing molecules as graphs may be more effective than those relying only on SMILES strings, as SMILES strings have limitations such as non-uniqueness \cite{https://doi.org/10.1002/wcms.1603}. Additionally, LLM-generated SMILES strings may not follow proper SMILES grammar, which renders them invalid and possibly constitute hallucinations. \textcolor{black}{In our work, datasets involving protein amino acid sequences perform similarly as those involving SMILES strings relative to SOTA models (\cref{tab:task_type_results}), which is less intuitive because proteins are naturally represented as sequences. One consideration is that SOTA models often encode evolutionary or structural information that is not learned by pretraining a LLM on natural language text. For example, BiComp-DTA uses evolutionary information from combining alignment-free and alignment-based methods in its protein encoding to predict drug-target dissociation constants, and Kinnings \etal ~\cite{BindingDB_ic50_ref} perform structure-based docking calculations to predict drug-target IC$_{50}$s.}

\textbf{Evidence of data contamination affecting \ourmodel performance was not observed} \textcolor{black}{We analyzed the percent overlap between TDC datasets and the PaLM-2 training data in \cref{tab:percent_contamination}. The large majority of datasets do not have any overlap with the PaLM-2 training data, while 7 datasets have some overlap. For these 7 datasets, we also filter the test set to remove the overlapping datapoints, and the performances on the filtered test sets do not decrease relative to the unfiltered test sets. We caution that our contamination analysis looks for direct character overlaps and thus may miss overlapping features that are represented in different forms between the PaLM-2 training data and TDC datasets. In particular, molecules are represented as SMILES strings in TDC but likely referred to by name in natural language text used to train a generic LLM.}

\subsection{Evidence of positive transfer across datasets with diverse drug types}

\begin{figure}
    \centering
    \includegraphics[width=0.9\textwidth]{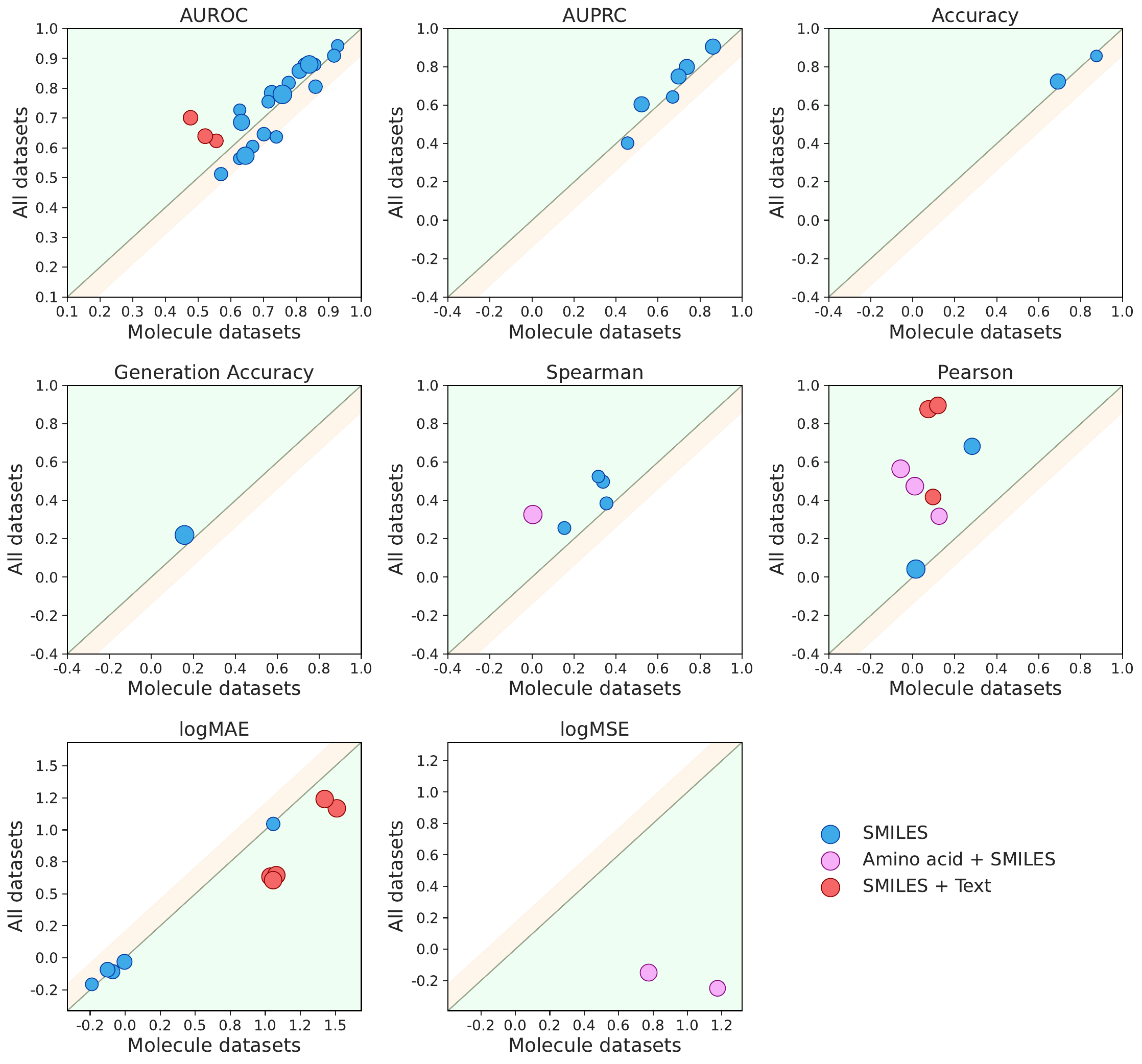}
    \vspace{6pt}
    \caption{\textbf{\ourmodel shows evidence of positive transfer across datasets with diverse drug types.} Performance of \ourmodels finetuned and evaluated on small molecule datasets. ``All datasets'' indicates a \ourmodels model finetuned on all TDC datasets, and ``Molecule datasets'' indicates a \ourmodels model finetuned on datasets containing molecules (datasets involving other drug types such as proteins or nucleic acids are not included in training). Datasets are colored by their feature types indicated in the legend, and marker sizes illustrate the number of data points in the task on a log scale. The larger shaded area in green indicates where ``All datasets'' is better than ``Molecule dataset'' (showing evidence of positive transfer), while the narrower orange shaded area indicates where the performance of ``Molecule datasets'' is near the performance of ``All dataset'' (defined as within 10\%). MAE and MSE values are log-transformed because the magnitudes of these values depend on the units of the outputs. Generation accuracy is the fraction of correct SMILES strings in the USPTO generation dataset.}
    \label{fig:drug_only}
\end{figure}

To assess positive transfer in \ourmodel, we trained a \ourmodels model only on TDC datasets containing molecules (excluding datasets involving other drug types such as proteins and nucleic acids) and compared the performance on small molecule datasets against the \ourmodels trained on all TDC datasets. The results in \cref{fig:drug_only} illustrate that the model trained on all datasets performs better than the model trained on small molecule datasets when evaluated on 43 out of 56 small molecule datasets.

To determine whether this improved performance was statistically significant, we performed a Wilcoxon signed-rank test on the small molecule dataset performances from both models (see Methods). The difference in performances between the model trained on all datasets and the model trained on small molecule datasets was highly significant ($\rho=1.4\times10^{-5}$), showing evidence of positive transfer between datasets involving diverse drug types and datasets involving small molecules. This is interesting given that other drug types such as proteins and nucleic acids are represented using their sequences, which are quite different from the SMILES strings used to represent small molecules. This may be because small molecules and proteins are more similar to each other than to natural language, so additional training steps away from PaLM-2 are ultimately helpful.

A model trained only on datasets in the ADMET benchmark group, which was a subset of small molecule datasets, was also evaluated in \cref{tab:drug_only_binary,tab:drug_only_regression_generation} in order to study positive transfer between small molecule datasets and ADMET datasets. For datasets in the ADMET benchmark, the model trained on small molecule datasets was the best model for only 4 ADMET datasets, while the model trained on ADMET datasets was the best model for 7 ADMET datasets. Interestingly, the model trained on all datasets was the best model for 11 ADMET datasets, suggesting that positive transfer between datasets with diverse drug types may be more impactful than positive transfer within small molecule datasets. This may support the use of a generalist model that expresses many drug types with the same string representation, rather than using separate representations for each drug type.

\subsection{Ablations}

\textbf{Model size and finetuning} We compared performances on TDC datasets for models of multiple sizes (\ourmodels and \ourmodelm) to examine the effect of model scale in \cref{fig:ablations_viz_1,fig:ablations_viz_2,tab:binary_results_sizes,tab:regression_generation_results_sizes}. Additionally, we evaluated \basemodels and \basemodelm as baseline generalist models in order to study the effect of biochemical domain finetuning. \ourmodelm outperformed \ourmodels on 57 out of 66 TDC datasets, suggesting that scale is beneficial within our tested sizes ($\rho=1.65\times10^{-7}$, Wilcoxon signed-rank test). Domain finetuning is also significant, as \ourmodel outperforms PaLM-2 on 60 datasets for the S models ($\rho=1.86\times10^{-10}$, Wilcoxon signed-rank test) and on 63 datasets for the M models ($\rho=3.58\times10^{-11}$, Wilcoxon signed-rank test).

\textbf{Number of shots and shot selection} We compared varying the number of shots in the prompt as well as whether the shots are selected from random datapoints or nearest neighbor datapoints in order to study the extent to which \ourmodel could exploit in-context learning in \cref{fig:ablations_viz_1,fig:ablations_viz_2,tab:binary_results_shots,tab:regression_generation_shots}. We observed that the best performances were not clearly skewed toward a particular prompting strategy, and pairwise comparisons did not yield statistically significant differences ($\rho > 0.05$, Wilcoxon signed-rank test). This is consistent with previous observations that zero-shot and few-shot prompting yielded marginal differences in performance when evaluated on a single task \cite{longpre2023flan}.

\textbf{Context removal} We constructed a unique context for each TDC dataset that provides background information for \ourmodel. We studied the effect of removing this context in \cref{tab:binary_results_nocontext,tab:regression_generation_nocontext}. We observed that removing the context reduced performance in 49 out of 66 datasets, which was also statistically significant ($\rho=4.9\times10^{-6}$, Wilcoxon signed-rank test). Given that we trained a single set of weights for all TDC datasets, this suggests that providing context may be a useful way to allow generalist LLMs to become effective predictors for specific tasks. This was especially true for the ToxCast dataset, which contained numerous subtasks corresponding to predicting toxicity in various assays. Each ToxCast assay draws from the same set of small molecules and measures toxicity, but the toxicity label differs based on the particular assay used. Without providing assay-specific information in the context, the model would have no way of differentiating subtasks with different labels.

\section{Discussion}
\label{sec:discussion}

To the best of our knowledge, \ourmodel is the first LLM trained on a wide variety of TDC datasets including small molecules, proteins, nucleic acids, cells, and diseases all in a single model. Interestingly, we found that including datasets without small molecules in our training, such as those only using proteins, enhances performance on small molecule datasets compared to models trained only on small molecule datasets. A previous generalist AI model finetuned for the biomedical domain (Med-PaLM M \cite{tu2023generalist}) observed positive transfer between chest X-ray report generation and chest X-ray classification tasks. There, the positive transfer was between tasks that were relatively closely related, as both tasks involved chest X-rays. However, the positive transfer observed in \ourmodel is between molecular tasks using SMILES strings and protein tasks using amino acid sequences, which are quite different.

\textcolor{black}{Although LLMs have demonstrated strong general language and reasoning capabilities, their effectiveness has been notably limited when it comes to tasks requiring specialized knowledge in chemistry~\cite{white_digital_discovery,ouyang2022training}}. For example, it was found that InstructGPT~\cite{ouyang2022training} rarely produced invalid SMILES strings, possibly suggesting that SMILES strings were encountered during training, but InstructGPT was unable to correctly predict the molecule name from the SMILES string. Thus, while non-finetuned LLMs may be able to capture grammar, understanding deeper connections is difficult, and domain finetuning may be necessary to achieve strong performance for therapeutics.

LLMs have previously been challenged by mathematical problems and regression, as it has been observed that smaller models have underperformed on mathematical benchmarks, which may be related to lossy compression of numbers in the model hidden states \cite{zhu2024language}. However, \ourmodel is often effective at performing regression, exceeding SOTA or achieving near-SOTA performance on \nregressionatleastnearsota out of 29 regression datasets. In other work, Gruver~\etal~\cite{gruver2023large} found that LLMs could be used for time series prediction, which was attributed to the LLMs' proclivity for periodicity often found in time series. LLMs have also been implemented as universal regressors \cite{song2024omnipred}, and LLMs finetuned for the chemistry domain have been used to perform regression tasks on molecules \cite{yu2024llasmol}. Taken together, these suggest that applying LLMs to regression problems may be a promising area to explore in multiple domains.

Since \ourmodel is a generalist model trained on a wide variety of tasks, we propose that it can \textcolor{black}{have a future role for end-to-end therapeutic development} spanning early-stage development such as target discovery to late-stage development such as clinical trial approval. \cref{fig:potential_use} illustrates an example for the case of developing small molecule drugs against type 2 diabetes. Here, \ourmodel \textcolor{black}{can be effective at identifying} genes associated with type 2 diabetes, predicting binding affinities of many small molecules against the protein target, predicting toxicities of the selected small molecules, and finally predicting the probability of clinical trial approval. \ourmodel can also be used for other tasks that are not illustrated, such as predicting drug permeability and drug synthesis reactions. 

However, at this stage, \ourmodel is in the research stage with scope for further improvement as the model is not an effective predictor for every task. In particular, the integration of the Gemini family of models \cite{geminiteam2024gemini} with \ourmodel remains an interesting possibility to enhance performance. Experimental validation and subsequent screening steps also remain an essential, complementary part of the therapeutic development pipeline. Overall, we believe the methodology behind \ourmodel (including the creation of \ourdata and the development of fine-tuned LLMs) represents a promising step towards using AI to contextualize and enhance many aspects of therapeutic development in future.

One important limitation of our work is that \ourmodel is not instruction-tuned to follow natural language because we were primarily interested in the accuracy of its predictions and did not want to limit this ability by also constraining \ourmodel to follow natural language. Therefore, unlike LLMs finetuned for medical question answering \cite{singhal_large_2023,singhal2023expertlevel,tu2023generalist}, \ourmodel is unable to explain its predictions to the user. This limits the benefits gained by training over a wide variety of tasks and may be an interesting area for future work.

\textcolor{black}{Additionally, it is important to note that LLMs are trained on large datasets, which increase the potential for data contamination and may result in overestimating their generalization. For \ourmodel, our data contamination analysis suggests that there is little overlap between the PaLM-2 training data and our evaluation data, and filtering the overlapping datapoints does not result in decreased performance. Nonetheless, our analysis does not account for different formats between the PaLM-2 training data and TDC data (e.g. using molecule names vs SMILES strings), and prospectively analyzing the performance of \ourmodel over time may be another avenue to study the effect of data contamination.}
\section{Conclusion}
\label{sec:conclusion}
Therapeutic development is an expensive process that involves many potential types of therapeutics as well as many criteria for approval. AI models may be useful tools for reducing the failure rate of therapeutic development by providing initial screening for multiple aspects of the development pipeline. Although further development and validation is required, we believe \ourmodel represents a notable advance towards a single generalist AI that can contextualize and aid many aspects of development ranging from target discovery to manufacturing.

\vspace{12pt}
\subsubsection*{Acknowledgments}
This project was a collaboration between teams at Google Research and Google DeepMind. We thank David~Belanger for the feedback and insight which significantly contributed to the enhancement of this report. We also thank Sami Lachgar, Lauren Winer, Maggie Shiels, Jessica Valdez, Jane Park, Jon Small, Aaron Abood, Rishad Patel, Uchechi Okereke, Annisah Um’rani, Alan Karthikesalingam, Anil Palepu, and Juraj Gottweis for their valuable insights, technical support and feedback during our research. We are also grateful to Zoubin Ghahramani, Raia Hadsell, Jon Shlens and Joelle Barral for their support during the course of this project.

\subsubsection*{Data availability}
Datasets utilized for developing, benchmarking, and evaluation of \ourmodel are publicly accessible with appropriate permissions. The Therapeutics Data Commons (TDC) datasets are accessible via their \href{https://tdcommons.ai/}{website}.

\vspace{12pt}
\subsubsection*{Competing interests}
This study was funded by Alphabet Inc and/or a subsidiary thereof (‘Alphabet’). T. T., C. S., D. F., V. N., and S. A. are employees of Alphabet and may own stock as part of the standard compensation package. 

\newpage
\setlength\bibitemsep{3pt}
\balance
\clearpage
\printbibliography
\end{refsection}

\newpage
\begin{refsection}

\clearpage

\renewcommand{\thesection}{A.\arabic{section}}
\renewcommand{\thefigure}{A.\arabic{figure}}
\renewcommand{\thetable}{A.\arabic{table}}
\renewcommand{\theequation}{A.\arabic{equation}}

\setcounter{section}{0}
\setcounter{figure}{0}
\setcounter{table}{0}
\setcounter{equation}{0}
\setcounter{page}{1}

\noindent \textbf{\LARGE{Appendix}}\\

\normalfont

\section{Additional data details}
\label{subsec:additional_details}

\begin{table}[htbp]
\centering
\caption{Excluded TDC datasets and reasons for exclusion.}
\label{tab:excluded_results}
\begin{tabular}{l|p{13cm}}
\toprule
Dataset name & Reason for exclusion \\
\midrule
QM7b & Prediction of quantum properties is not closely related to therapeutic development. \\ 
QM8 & Prediction of quantum properties is not closely related to therapeutic development. \\ 
QM9 & Prediction of quantum properties is not closely related to therapeutic development. \\ 
IEDB Jespersen & Amount of data is small, and token prediction is more difficult to implement in a LLM than binary classification. \\ 
PDB Jespersen & Amount of data is small, and token prediction is more difficult to implement in a LLM than binary classification. \\ 
DrugBank DDI & Large number of possible labels is difficult to implement in a LLM. \\ 
TWOSIDES & Large number of possible labels is difficult to implement in a LLM. \\ 
USPTO Catalyst & Large number of possible labels is difficult to implement in a LLM. \\ 
MOSES & No clear metric. \\ 
ZINC & No clear metric. \\ 
ChEMBL & No clear metric. \\ 
USPTO 50K & Subset of USPTO. \\ 
USPTO Reaction & Same dataset as USPTO. \\ 
\bottomrule
\end{tabular}
\end{table}

\begin{table}[htbp]
\centering
\caption{Hyperparameters used for \ourmodel finetuning.}
\label{tab:hyperparameters}
\begin{tabular}{lcc}
\toprule
Hyperparameter & \ourmodels & \ourmodelm \\
\midrule
Learning rate & $3\times10^{-5}$ & $1\times10^{-4}$ \\
Dropout rate & 0.05 & 0.15 \\
Batch size & 256 & 256 \\ 
Max token input length & 2048 & 2048 \\ 
Max token output length & 512 & 512 \\ 
\bottomrule
\end{tabular}
\end{table}

\begin{table}[htbp]
\centering
\caption{Example of prompts for binary classification datasets.}
\label{tab:example_prompts_binary}
\begin{tabular}{p{16cm}}
\toprule

\color{black}
\textbf{Instructions}: Answer the following question about drug properties.

\textbf{Context}: As a membrane separating circulating blood and brain extracellular fluid, the blood-brain barrier (BBB) is the protection layer that blocks most foreign drugs. Thus the ability of a drug to penetrate the barrier to deliver to the site of action forms a crucial challenge in development of drugs for central nervous system.

\textbf{Question}: Given a drug SMILES string, predict whether it

(A) does not cross the BBB (B) crosses the BBB

Drug SMILES: CN1C(=O)CN=C(C2=CCCCC2)c2cc(Cl)ccc21

\textbf{Answer: (B)}\\\\

\color{black}
\textbf{Instructions}: Answer the following question about peptide-MHC binding.

\textbf{Context}: In the human body, T cells monitor the existing peptides and trigger an immune response if the peptide is foreign. To decide whether or not if the peptide is not foreign, the peptide must bind to a major histocompatibility complex (MHC) molecule. Therefore, predicting peptide-MHC binding affinity is pivotal for determining immunogenicity. In some experiments, the peptide binding is measured against cells that express multiple MHCs, so the peptide could be binding any one of the possible MHCs. Class 1 MHC molecules bind to peptides that are usually 8-14 amino acids long and activate CD8 T cells.

\textbf{Question}: Given the amino acid sequence of the peptide and possible pseudo amino acid sequences of MHC 1, predict whether the peptide

(A) does not bind to any of the MHCs (B) binds to any of the MHCs

Peptide amino acid sequence: QLADETLLKV

Possible MHC pseudosequences: YFAMYGEKVAHTHVDTLYVRYHYYTWAEWAYTWY

\textbf{Answer: (B)}\\\\

\color{black}
\textbf{Instructions}: Answer the following question about miRNA protein interactions.

\textbf{Context}: MicroRNAs (miRNAs) are, small non-coding RNAs with 18–25 nucleotides, which are central regulators at the post-transcriptional level in both animals and plants. Perfect or near-perfect complementary binding of miRNAs and their target mRNA negatively regulates gene expression by accelerating mRNA degradation or suppressing mRNA translation.

\textbf{Question}: Given the miRNA mature sequence and target amino acid sequence, predict whether

(A) the miRNA and target do not interact (B) the miRNA and target interact

miRNA sequence: UUCCUGUCAGCCGUGGGUGCC

Target amino acid sequence: MSVNMDELRHQVMINQFVLAAGCAADQAKQLLQAAHWQFETALSTFF QETNIPNSHHHHQMMCTPSNTPATPPNFPDALAMFSKLRASEGLQSSNSPMTAAACSPPANFSPFWASSPPSHQAPWIPPSSPTTFHHLHRPQPTWPPGAQQGGAQQKAMAAMDGQR

\textbf{Answer: (A)}\\\\

\color{black}
\textbf{Instructions}: Answer the following question about clinical trials.

\textbf{Context}: Clinical trial is the most time and cost-consuming step in the drug discovery process. Phase 1 clinical trials test the safety and basic properties of a new drug or treatment in a small group of people for the first time. Optimizing and designing trials with machine learning could drastically lead to the speedup of delivery of life-saving therapeutics to patients. Clinical trial outcome prediction is a machine learning task that aims to forecast the outcome of clinical trials, such as the approval rate of a drug or treatment. It utilizes various clinical trial features, including the drug's molecular structure and patient disease.

\textbf{Question}: Given a drug SMILES string and disease, predict if the phase 1 trial

(A) would not be approved (B) would be approved

Drug SMILES: COC1=NC(N)=NC2=C1N=CN2[C@@H]1O[C@H](CO)[C@@H](O)[C@@H]1O

Disease: Chronic myeloproliferative disease

\textbf{Answer: (A)}

\color{black}

\hrule

\end{tabular}
\end{table}

\begin{table}[htbp]
\centering
\caption{Example of prompts for regression and generation datasets.}
\label{tab:example_prompts_regression_generation}
\begin{tabular}{p{16cm}}
\toprule

\color{black}
\textbf{Instructions}: Answer the following question about drug properties.

\textbf{Context}: The human colon epithelial cancer cell line, Caco-2, is used as an in vitro model to simulate the human intestinal tissue. The experimental result on the rate of drug passing through the Caco-2 cells can approximate the rate at which the drug permeates through the human intestinal tissue.

\textbf{Question}: Given a drug SMILES string, predict its normalized Caco-2 cell effective permeability from 000 to 1000, where 000 is minimum permeability and 1000 is maximum permeability.

Drug SMILES: O=C(O)COC(=O)Cc1ccccc1Nc1c(Cl)cccc1Cl

\textbf{Answer: 788}\\\\

\color{black}
\textbf{Instructions}: Answer the following question about drug responses.

\textbf{Context}: The same drug compound could have various levels of responses in different patients. To design drug for individual or a group with certain characteristics is the central goal of precision medicine. In experiments, IC50s of drugs were measured against cancer cell lines.

\textbf{Question}: Given a drug SMILES string and a cell line description, predict the normalized drug sensitivity from 000 to 1000, where 000 is minimum drug sensitivity and 1000 is maximum drug sensitivity.

Drug SMILES: CN1C=C(C2=CC=CC=C21)/C=C\textbackslash3/C4=C(C=CC=N4)NC3=O

Cell line description: SNU-1, stomach cell sourced from cancer

\textbf{Answer: 615}\\\\

\color{black}

\color{black}
\textbf{Instructions}: Answer the following question about drug target interactions.

\textbf{Context}: Drug-target binding is the physical interaction between a drug and a specific biological molecule, such as a protein or enzyme. This interaction is essential for the drug to exert its pharmacological effect. The strength of the drug-target binding is determined by the binding affinity, which is a measure of how tightly the drug binds to the target. Kd is the dissociation constant of a drug-target complex. It is the concentration of drug at which half of the drug-target complexes have dissociated. A lower Kd value indicates a stronger binding affinity.

\textbf{Question}: Given the target amino acid sequence and compound SMILES string, predict their normalized binding affinity Kd from 000 to 1000, where 000 is minimum Kd and 1000 is maximum Kd.

Drug SMILES: O=S(=O)(O)c1cccc2cccc(Nc3ccccc3)c12

Target amino acid sequence: MATVQQLEGRWRLVDSKGFDEYMKELGVGIALRKMGAMAKPDC IITCDGKNLTIKTESTLKTTQFSCTLGEKFEETTADGRKTQTVCNFTDGALVQHQEWDGKESTITRKLKDGKLVVECVMNNVTCTRIYEKVE

\textbf{Answer: 397}\\\\

\color{black}
\textbf{Instructions}: Answer the following question about reactions.

\textbf{Context}: Retrosynthesis is the process of finding a set of reactants that can synthesize a target molecule, i.e., product, which is a fundamental task in drug manufacturing. The target is recursively transformed into simpler precursor molecules until commercially available "starting" molecules are identified. In a data sample, there is only one product molecule, reactants can be one or multiple molecules.

\textbf{Question}: Given a product SMILES string, predict the reactant SMILES string.

Product SMILES: [CH2:12]1[C:7]2([CH2:6][CH2:5][O:15][CH2:1][CH2:8]2)[CH2:13][CH2:14][O:10][C:11]1=[O:17]

\textbf{Answer: [CH:1]12B[CH:5]([CH2:6][CH2:7][CH2:8]1)CCC2.[O:10]1[CH2:14][CH2:13][CH2:12] [CH2:11]1.[OH-:15].[Na+].[OH:17]O.Cl}\\\\

\hrule

\end{tabular}
\end{table}

\begin{table}[htbp]
\centering
\caption{Example of a 10-shot prompt for a binary classification dataset.}
\label{tab:example_prompts_fewshot}
\begin{tabular}{p{16cm}}
\toprule

\color{black}
\textbf{Instructions}: Answer the following question about drug properties.

\textbf{Context}: As a membrane separating circulating blood and brain extracellular fluid, the blood-brain barrier (BBB) is the protection layer that blocks most foreign drugs. Thus the ability of a drug to penetrate the barrier to deliver to the site of action forms a crucial challenge in development of drugs for central nervous system.

\textbf{Question}: Given a drug SMILES string, predict whether it

(A) does not cross the BBB (B) crosses the BBB \\

\color{black}
Drug SMILES: CN1C(=O)CN=C(c2ccccc2)c2cc(Cl)ccc21

Answer: (B)

Drug SMILES: CN1C(=O)CN=C(c2ccccc2F)c2cc(Cl)ccc21

Answer: (B)

Drug SMILES: CN1C(=S)CN=C(c2ccccc2)c2cc(Cl)ccc21

Answer: (B)

Drug SMILES: CP(C)(=O)CN1C(=O)CN=C(c2ccccc2)c2cc(Cl)ccc21

Answer: (B)

Drug SMILES: CN1C(=O)CN=C(c2ccccc2)c2cc([N+](=O)[O-])ccc21

Answer: (B)

Drug SMILES: CCN(CC)CCN1C(=O)CN=C(c2ccccc2F)c2cc(Cl)ccc21

Answer: (B)

Drug SMILES: O=C1CN=C(c2ccccc2)c2cc(Cl)ccc2N1CC1CC1

Answer: (B)

Drug SMILES: C\#CCN1C(=O)CN=C(c2ccccc2)c2cc(Cl)ccc21

Answer: (B)

Drug SMILES: O=C1CN=C(c2ccccc2)c2cc(Cl)ccc2N1CC(F)(F)F

Answer: (B)

Drug SMILES: CCS(=O)(=O)CCN1C(=O)CN=C(c2ccccc2F)c2cc(Cl)ccc21

Answer: (B)\\\\

\color{black}
Drug SMILES: CN1C(=O)CN=C(C2=CCCCC2)c2cc(Cl)ccc21

\color{black}
\textbf{Answer: (B)}\\

\color{black}
\hrule

\end{tabular}
\end{table}

\begin{figure}
    \centering
    \includegraphics[width=0.5\textwidth]{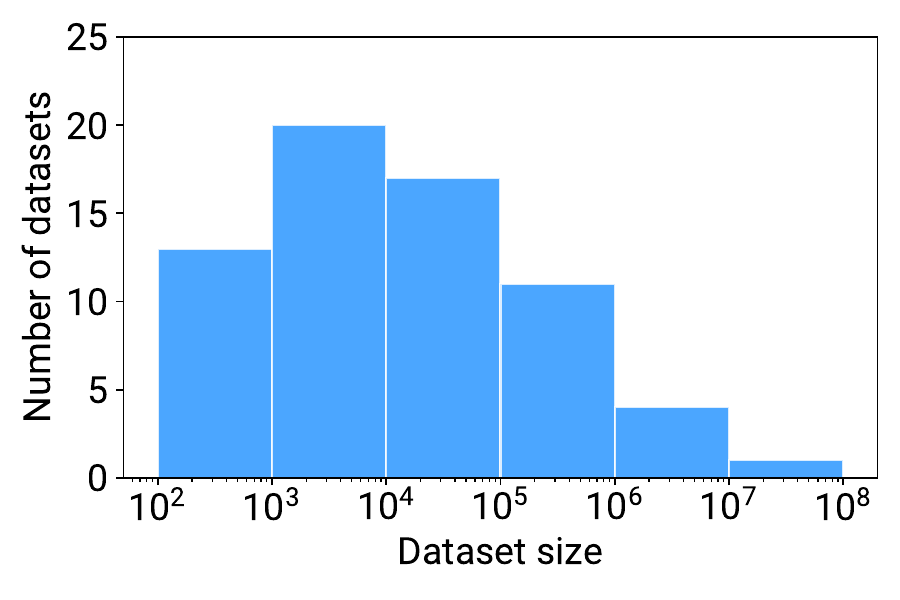}
    \caption{Distribution of TDC dataset sizes, aggregated over train, validation, and test sets. For datasets containing multiple subtasks, such as ToxCast which contains data for more than 600 different assays, the dataset size is calculated by summing over the sizes for each subtask.}
    \label{fig:dataset_size}
\end{figure}

\begin{figure}
    \centering
    \includegraphics[width=0.5\textwidth]{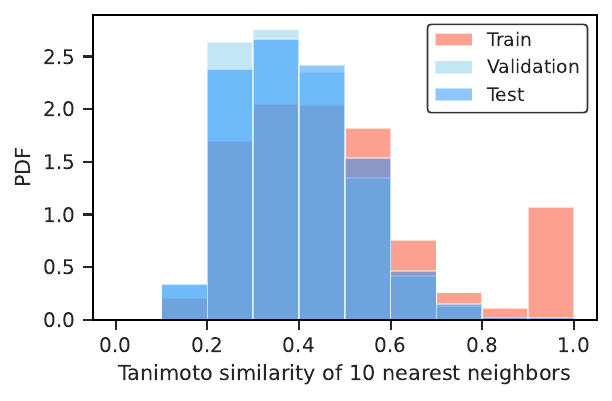}
    \caption{Distribution of the Tanimoto similarities for the 10 nearest neighbors in the AMES dataset. Nearest neighbors are calculated from the training set for training and validation sets, and from both the training and validation sets for the test set.}
    \label{fig:nn_distribution}
\end{figure}

\newpage

\section{Additional results}
\begin{table}[htbp]
\centering
\caption{Median relative difference of \ourmodelm performance from SOTA for datasets grouped by feature type. The relative difference is defined as (\ourmodel performance - SOTA) / SOTA. The signs were reversed for MAE and MSE metrics because lower MAE and MSE values correspond to better performances.}
\label{tab:task_type_results}
\begin{tabular}{lccccc}
\toprule
Feature type & Median relative difference of \ourmodel performance from SOTA \\
\midrule
SMILES + Text & 0.048 \\ 
Nucleotide + Amino acid & -0.007 \\ 
Amino acid & -0.080 \\ 
SMILES & -0.082 \\ 
Amino acid + SMILES & -0.482 \\ 
Nucleotide & -0.888 \\
\bottomrule
\end{tabular}
\end{table}

\newcolumntype{P}[1]{>{\centering\arraybackslash}p{#1}}
\begin{table}[htbp]
\centering
\caption{\ourmodelm performance compared with SOTA for each binary classification dataset, along with the feature types and metric type. Performances that are better than SOTA are bolded.}
\label{tab:binary_results}
\centerline{
\begin{tabular}{lP{2cm}P{2cm}P{1.5cm}P{1.5cm}P{1.5cm}}
\toprule
Dataset name & Feature type & Split method & Metric & SOTA & \ourmodel \\
\midrule
PAMPA NCATS & SMILES & Scaffold & AUROC & 0.900 \cite{PAMPA_NCATS_ref} & 0.668 \\ 
HIA Hou & SMILES & Scaffold & AUROC & 0.988 \cite{Caco2_Wang_ref} & \textbf{0.990} \\ 
Pgp Broccatelli & SMILES & Scaffold & AUROC & 0.935 \cite{Pgp_Broccatelli_ref} & \textbf{0.939} \\ 
Bioavailability Ma & SMILES & Scaffold & AUROC & 0.748 \cite{Bioavailability_Ma_ref} & 0.702 \\ 
BBB Martins & SMILES & Scaffold & AUROC & 0.915 \cite{BBB_Martins_ref} & 0.882 \\ 
CYP2C19 Veith & SMILES & Scaffold & AUROC & 0.890 \cite{CYP2C19_Veith_ref} & \textbf{0.895} \\ 
CYP2D6 Veith & SMILES & Scaffold & AUPRC & 0.739 \cite{CYP2D6_Veith_ref} & 0.659 \\ 
CYP3A4 Veith & SMILES & Scaffold & AUPRC & 0.904 \cite{CYP2D6_Veith_ref} & 0.840 \\ 
CYP1A2 Veith & SMILES & Scaffold & AUPRC & 0.900 \cite{CYP1A2_Veith_ref} & \textbf{0.914} \\ 
CYP2C9 Veith & SMILES & Scaffold & AUPRC & 0.839 \cite{CYP2D6_Veith_ref} & 0.788 \\ 
CYP2C9 Substrate CarbonMangels & SMILES & Scaffold & AUPRC & 0.441 \cite{CYP2C9_Substrate_CarbonMangels_ref} & 0.436 \\ 
CYP2D6 Substrate CarbonMangels & SMILES & Scaffold & AUPRC & 0.736 \cite{CYP2D6_Veith_ref} & 0.600 \\ 
CYP3A4 Substrate CarbonMangels & SMILES & Scaffold & AUROC & 0.662 \cite{CYP3A4_Substrate_CarbonMangels_ref} & 0.647 \\ 
hERG & SMILES & Scaffold & AUROC & 0.874 \cite{Bioavailability_Ma_ref} & \textbf{0.909} \\ 
AMES & SMILES & Scaffold & AUROC & 0.871 \cite{Pgp_Broccatelli_ref} & 0.786 \\ 
DILI & SMILES & Scaffold & AUROC & 0.925 \cite{Pgp_Broccatelli_ref} & 0.882 \\ 
Skin Reaction & SMILES & Scaffold & AUROC & 0.840 \cite{Skin_Reaction_ref} & 0.615 \\ 
Carcinogens Lagunin & SMILES & Scaffold & Accuracy & 0.770 \cite{Carcinogens_Lagunin_ref} & \textbf{0.786} \\ 
Tox21 & SMILES & Scaffold & AUROC & 0.961 \cite{Tox21_ref} & 0.882 \\ 
ClinTox & SMILES & Scaffold & AUROC & 0.948 \cite{ClinTox_ref} & 0.863 \\ 
herg central & SMILES & Scaffold & AUROC & 0.860 \cite{herg_central_ref} & \textbf{0.888} \\ 
hERG Karim & SMILES & Scaffold & Accuracy & 0.770 \cite{hERG_Karim_ref} & 0.745 \\ 
ToxCast & SMILES & Scaffold & AUROC & 0.777 \cite{ClinTox_ref} & \textbf{0.792} \\ 
SARSCoV2 Vitro Touret & SMILES & Scaffold & AUROC & 0.640 \cite{SARSCoV2_Vitro_Touret_ref} & 0.601 \\ 
SARSCOV2 3CLPro Diamond & SMILES & Scaffold & AUROC & 0.800 \cite{SARSCOV2_3CLPro_Diamond_ref} & 0.712 \\ 
HIV & SMILES & Scaffold & AUROC & 0.851 \cite{HIV_ref} & 0.732 \\ 
SAbDab Chen & Amino acid & Random & AUPRC & 0.510 \cite{SAbDab_Chen_ref} & 0.473 \\ 
HuRI & Amino acid & Cold-start & AUPRC & 0.724 \cite{HuRI_ref} & \textbf{0.753} \\ 
miRTarBase & Nucleotide + Amino acid & Random & Accuracy & 0.804 \cite{miRTarBase_ref} & 0.799 \\ 
MHC1 IEDB IMGT Nielsen & Amino acid & Random & AUROC & 0.986 \cite{MHC1_IEDB_IMGT_Nielsen_ref} & 0.907 \\ 
MHC2 IEDB Jensen & Amino acid & Random & AUROC & 0.940 \cite{MHC2_IEDB_Jensen_ref} & 0.863 \\ 
weber & Amino acid & Cold-start & AUROC & 0.870 \cite{weber_ref} & 0.743 \\ 
phase1 & SMILES + Text & Cold-start & AUROC & 0.576 \cite{phase1_ref} & \textbf{0.667} \\ 
phase2 & SMILES + Text & Cold-start & AUROC & 0.645 \cite{phase1_ref} & \textbf{0.676} \\ 
phase3 & SMILES + Text & Cold-start & AUROC & 0.723 \cite{phase1_ref} & \textbf{0.728} \\ 
butkiewicz & SMILES & Random & AUROC & 0.840 \cite{orexin1_receptor_butkiewicz_ref} & 0.566 \\
\bottomrule
\end{tabular}
}
\end{table}
\newcolumntype{P}[1]{>{\centering\arraybackslash}p{#1}}
\begin{table}[htbp]
\centering
\caption{\ourmodelm performance compared with SOTA for each regression and generation dataset, along with the feature types and metric type. Performances that are better than SOTA are bolded. Datasets for which we did not find a SOTA are marked as N/A.}
\label{tab:regression_generation_results}
\centerline{
\begin{tabular}{lP{2.5cm}P{1.8cm}P{2cm}P{1.8cm}P{1.8cm}P{1.8cm}}
\toprule
Dataset name & Feature type & Split method & Metric & SOTA & \ourmodel \\
\midrule
Caco2 Wang & SMILES & Scaffold & MAE & 0.285 \cite{Caco2_Wang_ref} & 0.432 \\ 
Lipophilicity AstraZeneca & SMILES & Scaffold & MAE & 0.467 \cite{Lipophilicity_AstraZeneca_ref} & 0.587 \\ 
Solubility AqSolDB & SMILES & Scaffold & MAE & 0.761 \cite{Lipophilicity_AstraZeneca_ref} & 0.987 \\ 
PPBR AZ & SMILES & Scaffold & MAE & 7.788 \cite{Lipophilicity_AstraZeneca_ref} & 9.108 \\ 
VDss Lombardo & SMILES & Scaffold & Spearman & 0.627 \cite{VDss_Lombardo_ref} & 0.609 \\ 
Half Life Obach & SMILES & Scaffold & Spearman & 0.547 \cite{Half_Life_Obach_ref} & 0.448 \\ 
Clearance Hepatocyte AZ & SMILES & Scaffold & Spearman & 0.440 \cite{Clearance_Hepatocyte_AZ_ref} & 0.385 \\ 
Clearance Microsome AZ & SMILES & Scaffold & Spearman & 0.625 \cite{Caco2_Wang_ref} & 0.413 \\ 
LD50 Zhu & SMILES & Scaffold & MAE & 0.552 \cite{LD50_Zhu_ref} & 0.618 \\ 
USPTO Yields & SMILES & Random & Pearson & 0.361 \cite{USPTO_Yields_ref} & 0.070 \\ 
Buchwald Hartwig & SMILES & Random & Pearson & 0.786 \cite{USPTO_Yields_ref} & \textbf{0.905} \\ 
TAP & Amino acid & Random & MAE & N/A & \textbf{4.983} \\ 
Leenay & Nucleotide & Random & Spearman & 0.740 \cite{Leenay_ref} & 0.083 \\ 
BindingDB kd & Amino acid + SMILES & Cold-start & Pearson & 0.712 \cite{BindingDB_kd_ref} & 0.391 \\ 
BindingDB ic50 & Amino acid + SMILES & Cold-start & Spearman & 0.637 \cite{BindingDB_ic50_ref} & 0.311 \\ 
BindingDB ki & Amino acid + SMILES & Cold-start & Pearson & 0.840 \cite{BindingDB_ki_ref} & 0.726 \\ 
BindingDB Patent & Amino acid + SMILES & Temporal & Pearson & 0.588 \cite{BindingDB_Patent_ref} & 0.531 \\ 
DAVIS & Amino acid + SMILES & Cold-start & MSE & 0.219 \cite{DAVIS_ref} & 0.704 \\ 
KIBA & Amino acid + SMILES & Cold-start & MSE & 0.154 \cite{DAVIS_ref} & 0.548 \\ 
DisGeNET & Amino acid + Text & Random & MAE & N/A & \textbf{0.057} \\ 
GDSC1 & SMILES + Text & Random & Pearson & 0.860 \cite{GDSC1_ref} & \textbf{0.887} \\ 
GDSC2 & SMILES + Text & Random & Pearson & 0.860 \cite{GDSC1_ref} & \textbf{0.900} \\ 
DrugComb CSS & SMILES + Text & Combination & MAE & 16.858 \cite{DrugComb_CSS_ref} & \textbf{14.057} \\ 
OncoPolyPharmacology & SMILES + Text & Combination & Pearson & 0.730 \cite{OncoPolyPharmacology_ref} & 0.552 \\ 
Protein SAbDab & Amino acid & Random & MAE & N/A & \textbf{1.268} \\ 
DrugComb HSA & SMILES + Text & Combination & MAE & 4.453 \cite{DrugComb_CSS_ref} & \textbf{4.118} \\ 
DrugComb Loewe & SMILES + Text & Combination & MAE & 9.184 \cite{DrugComb_CSS_ref} & 17.381 \\ 
DrugComb Bliss & SMILES + Text & Combination & MAE & 4.560 \cite{DrugComb_CSS_ref} & \textbf{4.104} \\ 
DrugComb ZIP & SMILES + Text & Combination & MAE & 4.027 \cite{DrugComb_CSS_ref} & \textbf{3.777} \\ 
USPTO & SMILES & Random & Generation Accuracy & 0.415 \cite{USPTO_ref} & 0.239 \\
\bottomrule
\end{tabular}
}
\end{table}

\newcolumntype{P}[1]{>{\centering\arraybackslash}p{#1}}
\begin{table}[htbp]
\centering
\caption{Performances on binary classification datasets for \basemodels, \basemodelm, \ourmodels and \ourmodelm. The best performances are bolded.}
\label{tab:binary_results_sizes}
\centerline{
\begin{tabular}{lccccc}
\toprule
Dataset name & Metric & \basemodels & \basemodelm & \ourmodels & \ourmodelm \\
\midrule
PAMPA NCATS & AUROC & 0.640 & 0.661 & 0.646 & \textbf{0.668} \\ 
HIA Hou & AUROC & 0.837 & 0.711 & 0.942 & \textbf{0.990} \\ 
Pgp Broccatelli & AUROC & 0.791 & 0.848 & 0.909 & \textbf{0.939} \\ 
Bioavailability Ma & AUROC & 0.492 & 0.564 & 0.605 & \textbf{0.702} \\ 
BBB Martins & AUROC & 0.732 & 0.616 & 0.805 & \textbf{0.882} \\ 
CYP2C19 Veith & AUROC & 0.608 & 0.627 & 0.877 & \textbf{0.895} \\ 
CYP2D6 Veith & AUPRC & 0.170 & 0.198 & 0.605 & \textbf{0.659} \\ 
CYP3A4 Veith & AUPRC & 0.544 & 0.544 & 0.800 & \textbf{0.840} \\ 
CYP1A2 Veith & AUPRC & 0.590 & 0.594 & 0.906 & \textbf{0.914} \\ 
CYP2C9 Veith & AUPRC & 0.336 & 0.401 & 0.750 & \textbf{0.788} \\ 
CYP2C9 Substrate CarbonMangels & AUPRC & 0.340 & 0.308 & 0.403 & \textbf{0.436} \\ 
CYP2D6 Substrate CarbonMangels & AUPRC & 0.380 & 0.575 & \textbf{0.643} & 0.600 \\ 
CYP3A4 Substrate CarbonMangels & AUROC & 0.560 & 0.604 & 0.637 & \textbf{0.647} \\ 
hERG & AUROC & 0.701 & 0.756 & 0.879 & \textbf{0.909} \\ 
AMES & AUROC & 0.614 & 0.563 & 0.785 & \textbf{0.786} \\ 
DILI & AUROC & 0.643 & 0.741 & 0.727 & \textbf{0.882} \\ 
Skin Reaction & AUROC & 0.431 & 0.553 & 0.564 & \textbf{0.615} \\ 
Carcinogens Lagunin & Accuracy & 0.821 & 0.714 & \textbf{0.857} & 0.786 \\ 
Tox21 & AUROC & 0.406 & 0.610 & 0.858 & \textbf{0.882} \\ 
ClinTox & AUROC & 0.387 & 0.471 & 0.818 & \textbf{0.863} \\ 
herg central & AUROC & 0.491 & 0.509 & 0.880 & \textbf{0.888} \\ 
hERG Karim & Accuracy & 0.570 & 0.555 & 0.724 & \textbf{0.745} \\ 
ToxCast & AUROC & 0.455 & 0.530 & 0.779 & \textbf{0.792} \\ 
SARSCoV2 Vitro Touret & AUROC & 0.580 & 0.556 & 0.512 & \textbf{0.601} \\ 
SARSCOV2 3CLPro Diamond & AUROC & 0.371 & 0.619 & \textbf{0.755} & 0.712 \\ 
HIV & AUROC & 0.436 & 0.494 & 0.686 & \textbf{0.732} \\ 
SAbDab Chen & AUPRC & 0.437 & \textbf{0.545} & 0.390 & 0.473 \\ 
HuRI & AUPRC & 0.509 & 0.501 & 0.705 & \textbf{0.753} \\ 
miRTarBase & Accuracy & 0.502 & 0.499 & 0.765 & \textbf{0.799} \\ 
MHC1 IEDB IMGT Nielsen & AUROC & 0.548 & 0.557 & \textbf{0.913} & 0.907 \\ 
MHC2 IEDB Jensen & AUROC & 0.617 & 0.604 & 0.781 & \textbf{0.863} \\ 
weber & AUROC & 0.666 & 0.680 & 0.738 & \textbf{0.743} \\ 
phase1 & AUROC & 0.524 & 0.490 & 0.624 & \textbf{0.667} \\ 
phase2 & AUROC & 0.529 & 0.519 & 0.639 & \textbf{0.676} \\ 
phase3 & AUROC & 0.527 & 0.508 & 0.701 & \textbf{0.728} \\ 
butkiewicz & AUROC & 0.504 & 0.499 & \textbf{0.574} & 0.566 \\
\bottomrule
\end{tabular}
}
\end{table}

\newcolumntype{P}[1]{>{\centering\arraybackslash}p{#1}}
\begin{table}[htbp]
\centering
\caption{Performances on regression and generation datasets for \basemodels, \basemodelm, \ourmodels and \ourmodelm. The best performances are bolded.}
\label{tab:regression_generation_results_sizes}
\centerline{
\begin{tabular}{lccccc}
\toprule
Dataset name & Metric & \basemodels & \basemodelm & \ourmodels & \ourmodelm \\
\midrule
Caco2 Wang & MAE & 0.457 & 1.680 & 0.621 & \textbf{0.432} \\ 
Lipophilicity AstraZeneca & MAE & 1.108 & 1.189 & 0.779 & \textbf{0.587} \\ 
Solubility AqSolDB & MAE & 2.536 & 5.427 & \textbf{0.931} & 0.987 \\ 
PPBR AZ & MAE & 13.104 & 32.447 & 11.138 & \textbf{9.108} \\ 
VDss Lombardo & Spearman & -0.062 & 0.174 & 0.497 & \textbf{0.609} \\ 
Half Life Obach & Spearman & -0.033 & 0.380 & \textbf{0.525} & 0.448 \\ 
Clearance Hepatocyte AZ & Spearman & 0.240 & 0.075 & 0.256 & \textbf{0.385} \\ 
Clearance Microsome AZ & Spearman & 0.337 & 0.024 & 0.385 & \textbf{0.413} \\ 
LD50 Zhu & MAE & 0.823 & 0.971 & 0.808 & \textbf{0.618} \\ 
USPTO Yields & Pearson & 0.001 & 0.002 & 0.042 & \textbf{0.070} \\ 
Buchwald Hartwig & Pearson & 0.333 & 0.089 & 0.682 & \textbf{0.905} \\ 
TAP & MAE & \textbf{2.480} & 3.536 & 5.075 & 4.983 \\ 
Leenay & Spearman & 0.004 & -0.010 & 0.048 & \textbf{0.083} \\ 
BindingDB kd & Pearson & 0.089 & 0.087 & 0.317 & \textbf{0.391} \\ 
BindingDB ic50 & Spearman & 0.033 & -0.114 & \textbf{0.326} & 0.311 \\ 
BindingDB ki & Pearson & 0.055 & 0.026 & 0.565 & \textbf{0.726} \\ 
BindingDB Patent & Pearson & -0.022 & 0.031 & 0.474 & \textbf{0.531} \\ 
DAVIS & MSE & 5.102 & 5.145 & \textbf{0.564} & 0.704 \\ 
KIBA & MSE & 8.530 & 8.777 & 0.709 & \textbf{0.548} \\ 
DisGeNET & MAE & 0.088 & 0.134 & 0.059 & \textbf{0.057} \\ 
GDSC1 & Pearson & -0.042 & 0.006 & 0.876 & \textbf{0.887} \\ 
GDSC2 & Pearson & -0.058 & 0.010 & 0.896 & \textbf{0.900} \\ 
DrugComb CSS & MAE & 27.159 & 24.66 & 14.740 & \textbf{14.057} \\ 
OncoPolyPharmacology & Pearson & 0.056 & 0.017 & 0.418 & \textbf{0.552} \\ 
Protein SAbDab & MAE & 1.282 & \textbf{1.236} & 1.432 & 1.268 \\ 
DrugComb HSA & MAE & 5.885 & 4.485 & 4.311 & \textbf{4.118} \\ 
DrugComb Loewe & MAE & \textbf{16.456} & 20.865 & 17.428 & 17.381 \\ 
DrugComb Bliss & MAE & 5.675 & 4.565 & 4.425 & \textbf{4.104} \\ 
DrugComb ZIP & MAE & 5.735 & 4.742 & 4.047 & \textbf{3.777} \\ 
USPTO & Generation Accuracy & 0.000 & 0.000 & 0.220 & \textbf{0.239} \\
\bottomrule
\end{tabular}
}
\end{table}
\newcolumntype{P}[1]{>{\centering\arraybackslash}p{#1}}

\begin{table}[htbp]
\centering
\caption{Performances on binary classification datasets for varying few-shot prompting strategies with \ourmodels. The number of shots is varied between 0, 1, 5, and 10, and the shots are either chosen randomly or based on the nearest neighbors (KNN). Nearest neighbors are determined by Tanimoto similarities for molecules and sequence identity for proteins and nucleotides. The best performances are bolded.}
\label{tab:binary_results_shots}
\centerline{
\begin{tabular}{lP{1.5cm}P{1.2cm}P{1.2cm}P{1.2cm}P{1.2cm}P{1.2cm}P{1.2cm}P{1.2cm}}
\toprule
Dataset name & Metric & 0-shot & 1-shot random & 5-shot random & 10-shot random & 1-shot KNN & 5-shot KNN & 10-shot KNN \\
\midrule
PAMPA NCATS & AUROC & \textbf{0.677} & 0.650 & 0.649 & 0.633 & 0.671 & 0.662 & 0.646 \\ 
HIA Hou & AUROC & 0.956 & \textbf{0.960} & 0.947 & 0.939 & 0.953 & 0.952 & 0.942 \\ 
Pgp Broccatelli & AUROC & 0.906 & 0.903 & 0.908 & 0.908 & 0.905 & 0.908 & \textbf{0.909} \\ 
Bioavailability Ma & AUROC & 0.607 & 0.616 & \textbf{0.619} & 0.597 & 0.611 & 0.606 & 0.605 \\ 
BBB Martins & AUROC & 0.807 & 0.806 & 0.800 & 0.799 & \textbf{0.811} & 0.808 & 0.805 \\ 
CYP2C19 Veith & AUROC & 0.874 & 0.875 & 0.876 & 0.875 & 0.875 & \textbf{0.877} & 0.877 \\ 
CYP2D6 Veith & AUPRC & 0.598 & 0.600 & 0.601 & 0.604 & 0.604 & 0.605 & \textbf{0.605} \\ 
CYP3A4 Veith & AUPRC & 0.801 & 0.802 & 0.802 & 0.799 & 0.803 & \textbf{0.803} & 0.800 \\ 
CYP1A2 Veith & AUPRC & 0.906 & \textbf{0.907} & 0.906 & 0.906 & 0.907 & 0.906 & 0.906 \\ 
CYP2C9 Veith & AUPRC & 0.751 & 0.748 & \textbf{0.752} & 0.750 & 0.748 & 0.751 & 0.750 \\ 
CYP2C9 Substrate CarbonMangels & AUPRC & 0.416 & 0.408 & \textbf{0.423} & 0.411 & 0.412 & 0.414 & 0.403 \\ 
CYP2D6 Substrate CarbonMangels & AUPRC & 0.624 & 0.634 & 0.624 & 0.624 & 0.637 & \textbf{0.648} & 0.643 \\ 
CYP3A4 Substrate CarbonMangels & AUROC & 0.645 & 0.645 & 0.640 & 0.639 & \textbf{0.646} & 0.641 & 0.637 \\ 
hERG & AUROC & 0.869 & 0.868 & 0.868 & 0.877 & 0.873 & 0.875 & \textbf{0.879} \\ 
AMES & AUROC & 0.780 & 0.784 & 0.781 & 0.783 & 0.781 & 0.783 & \textbf{0.785} \\ 
DILI & AUROC & 0.697 & 0.708 & 0.702 & 0.715 & 0.717 & 0.713 & \textbf{0.727} \\ 
Skin Reaction & AUROC & \textbf{0.572} & 0.566 & 0.544 & 0.549 & 0.570 & 0.552 & 0.564 \\ 
Carcinogens Lagunin & Accuracy & 0.839 & 0.857 & 0.875 & \textbf{0.893} & 0.857 & 0.857 & 0.857 \\ 
Tox21 & AUROC & 0.857 & 0.856 & 0.858 & 0.856 & 0.858 & \textbf{0.858} & 0.858 \\ 
ClinTox & AUROC & 0.801 & 0.796 & 0.800 & 0.814 & 0.811 & 0.807 & \textbf{0.818} \\ 
herg central & AUROC & 0.878 & 0.879 & 0.880 & \textbf{0.880} & 0.880 & 0.879 & 0.880 \\ 
hERG Karim & Accuracy & 0.711 & 0.709 & 0.714 & 0.715 & 0.714 & \textbf{0.725} & 0.724 \\ 
ToxCast & AUROC & 0.777 & 0.778 & 0.778 & 0.777 & 0.779 & 0.779 & \textbf{0.779} \\ 
SARSCoV2 Vitro Touret & AUROC & 0.525 & 0.526 & 0.519 & \textbf{0.542} & 0.521 & 0.525 & 0.512 \\ 
SARSCOV2 3CLPro Diamond & AUROC & 0.723 & 0.739 & 0.744 & 0.748 & 0.717 & 0.746 & \textbf{0.755} \\ 
HIV & AUROC & 0.675 & 0.683 & \textbf{0.691} & 0.686 & 0.675 & 0.686 & 0.686 \\ 
SAbDab Chen & AUPRC & 0.369 & 0.390 & 0.385 & 0.373 & \textbf{0.399} & 0.385 & 0.390 \\ 
HuRI & AUPRC & 0.698 & 0.705 & 0.705 & 0.705 & 0.705 & \textbf{0.705} & 0.705 \\ 
miRTarBase & Accuracy & 0.765 & \textbf{0.765} & 0.765 & 0.765 & 0.765 & 0.765 & 0.765 \\ 
MHC1 IEDB IMGT Nielsen & AUROC & 0.912 & 0.913 & 0.913 & 0.913 & \textbf{0.913} & 0.913 & 0.913 \\ 
MHC2 IEDB Jensen & AUROC & 0.775 & 0.780 & 0.780 & \textbf{0.783} & 0.778 & 0.778 & 0.781 \\ 
weber & AUROC & 0.737 & 0.738 & 0.738 & 0.738 & \textbf{0.738} & 0.738 & 0.738 \\ 
phase1 & AUROC & 0.628 & \textbf{0.634} & 0.627 & 0.628 & 0.631 & 0.628 & 0.624 \\ 
phase2 & AUROC & 0.641 & 0.642 & 0.640 & 0.638 & \textbf{0.642} & 0.640 & 0.639 \\ 
phase3 & AUROC & 0.693 & 0.696 & 0.697 & 0.692 & 0.695 & 0.697 & \textbf{0.701} \\ 
butkiewicz & AUROC & 0.581 & 0.548 & 0.582 & \textbf{0.593} & 0.558 & 0.538 & 0.574 \\
\bottomrule
\end{tabular}
}
\end{table}

\newcolumntype{P}[1]{>{\centering\arraybackslash}p{#1}}

\begin{table}[htbp]
\centering
\caption{Performances on regression and generation datasets for varying few-shot prompting strategies with \ourmodels. The number of shots is varied between 0, 1, 5, and 10, and the shots are either chosen randomly or based on the nearest neighbors (KNN). Nearest neighbors are determined by Tanimoto similarities for molecules and sequence identity for proteins and nucleotides. The best performances are bolded.}
\label{tab:regression_generation_shots}
\centerline{
\begin{tabular}{lP{1.5cm}P{1.2cm}P{1.2cm}P{1.2cm}P{1.2cm}P{1.2cm}P{1.2cm}P{1.2cm}}
\toprule
Dataset name & Metric & 0-shot & 1-shot random & 5-shot random & 10-shot random & 1-shot KNN & 5-shot KNN & 10-shot KNN \\
\midrule
Caco2 Wang & MAE & 0.605 & 0.621 & 0.627 & 0.613 & 0.616 & \textbf{0.597} & 0.621 \\ 
Lipophilicity AstraZeneca & MAE & 0.803 & 0.812 & 0.791 & 0.782 & 0.818 & 0.793 & \textbf{0.779} \\ 
Solubility AqSolDB & MAE & \textbf{0.899} & 0.918 & 0.915 & 0.921 & 0.919 & 0.917 & 0.931 \\ 
PPBR AZ & MAE & 10.814 & \textbf{10.727} & 10.854 & 11.024 & 10.935 & 11.129 & 11.138 \\ 
VDss Lombardo & Spearman & 0.496 & 0.496 & \textbf{0.508} & 0.487 & 0.486 & 0.488 & 0.497 \\ 
Half Life Obach & Spearman & 0.494 & 0.502 & 0.503 & 0.489 & 0.523 & 0.499 & \textbf{0.525} \\ 
Clearance Hepatocyte AZ & Spearman & 0.255 & 0.285 & 0.252 & 0.258 & 0.285 & \textbf{0.303} & 0.256 \\ 
Clearance Microsome AZ & Spearman & 0.401 & 0.406 & 0.401 & \textbf{0.406} & 0.403 & 0.386 & 0.385 \\ 
LD50 Zhu & MAE & 0.815 & 0.811 & 0.809 & 0.809 & 0.810 & \textbf{0.807} & 0.808 \\ 
USPTO Yields & Pearson & 0.010 & \textbf{0.044} & 0.040 & 0.038 & 0.044 & 0.042 & 0.042 \\ 
Buchwald Hartwig & Pearson & 0.736 & 0.802 & \textbf{0.809} & 0.491 & 0.800 & 0.808 & 0.682 \\ 
TAP & MAE & 5.150 & 5.092 & 5.137 & \textbf{5.029} & 5.087 & 5.075 & 5.075 \\ 
Leenay & Spearman & 0.036 & 0.035 & 0.034 & \textbf{0.063} & 0.032 & 0.055 & 0.048 \\ 
BindingDB kd & Pearson & 0.314 & 0.314 & \textbf{0.320} & 0.305 & 0.320 & 0.318 & 0.317 \\ 
BindingDB ic50 & Spearman & \textbf{0.331} & 0.327 & 0.327 & 0.327 & 0.327 & 0.326 & 0.326 \\ 
BindingDB ki & Pearson & 0.569 & 0.574 & 0.568 & 0.568 & \textbf{0.576} & 0.565 & 0.565 \\ 
BindingDB Patent & Pearson & \textbf{0.483} & 0.474 & 0.474 & 0.474 & 0.474 & 0.474 & 0.474 \\ 
DAVIS & MSE & \textbf{0.561} & 0.561 & 0.570 & 0.561 & 0.561 & 0.564 & 0.564 \\ 
KIBA & MSE & 0.743 & 0.711 & 0.715 & 0.718 & \textbf{0.707} & 0.709 & 0.709 \\ 
DisGeNET & MAE & 0.060 & \textbf{0.059} & 0.059 & 0.059 & 0.059 & 0.059 & 0.059 \\ 
GDSC1 & Pearson & \textbf{0.876} & 0.876 & 0.875 & 0.875 & 0.876 & 0.876 & 0.876 \\ 
GDSC2 & Pearson & 0.895 & 0.895 & 0.895 & 0.895 & 0.895 & 0.895 & \textbf{0.896} \\ 
DrugComb CSS & MAE & 14.779 & 14.775 & \textbf{14.729} & 14.749 & 14.781 & 14.743 & 14.740 \\ 
OncoPolyPharmacology & Pearson & 0.423 & 0.427 & 0.427 & 0.418 & \textbf{0.432} & 0.424 & 0.418 \\ 
Protein SAbDab & MAE & 1.399 & \textbf{1.384} & 1.427 & 1.406 & 1.398 & 1.420 & 1.432 \\ 
DrugComb HSA & MAE & 4.298 & 4.297 & 4.300 & 4.312 & \textbf{4.296} & 4.299 & 4.311 \\ 
DrugComb Loewe & MAE & 17.425 & 17.454 & 17.435 & \textbf{17.424} & 17.455 & 17.440 & 17.428 \\ 
DrugComb Bliss & MAE & \textbf{4.284} & 4.284 & 4.293 & 4.352 & 4.285 & 4.294 & 4.425 \\ 
DrugComb ZIP & MAE & \textbf{4.014} & 4.021 & 4.035 & 4.045 & 4.022 & 4.035 & 4.047 \\ 
USPTO & Generation Accuracy & \textbf{0.225} & 0.221 & 0.212 & 0.209 & 0.221 & 0.220 & 0.220 \\
\bottomrule
\end{tabular}
}
\end{table}

\newcolumntype{P}[1]{>{\centering\arraybackslash}p{#1}}

\begin{table}[htbp]
\centering
\caption{Performances on binary classification datasets with \ourmodels using 10-shot KNN prompting, with and without context. The best performances are bolded.}
\label{tab:binary_results_nocontext}
\centerline{
\begin{tabular}{lccc}
\toprule
Dataset name & Metric & 10-shot KNN no context & 10-shot KNN with context \\
\midrule
PAMPA NCATS & AUROC & \textbf{0.664} & 0.646 \\ 
HIA Hou & AUROC & 0.905 & \textbf{0.942} \\ 
Pgp Broccatelli & AUROC & 0.884 & \textbf{0.909} \\ 
Bioavailability Ma & AUROC & 0.592 & \textbf{0.605} \\ 
BBB Martins & AUROC & 0.797 & \textbf{0.805} \\ 
CYP2C19 Veith & AUROC & 0.875 & \textbf{0.877} \\ 
CYP2D6 Veith & AUPRC & 0.596 & \textbf{0.605} \\ 
CYP3A4 Veith & AUPRC & \textbf{0.805} & 0.800 \\ 
CYP1A2 Veith & AUPRC & 0.898 & \textbf{0.906} \\ 
CYP2C9 Veith & AUPRC & 0.750 & \textbf{0.750} \\ 
CYP2C9 Substrate CarbonMangels & AUPRC & 0.358 & \textbf{0.403} \\ 
CYP2D6 Substrate CarbonMangels & AUPRC & \textbf{0.645} & 0.643 \\ 
CYP3A4 Substrate CarbonMangels & AUROC & \textbf{0.639} & 0.637 \\ 
hERG & AUROC & 0.868 & \textbf{0.879} \\ 
AMES & AUROC & 0.747 & \textbf{0.785} \\ 
DILI & AUROC & 0.633 & \textbf{0.727} \\ 
Skin Reaction & AUROC & 0.529 & \textbf{0.564} \\ 
Carcinogens Lagunin & Accuracy & \textbf{0.857} & 0.857 \\ 
Tox21 & AUROC & 0.828 & \textbf{0.858} \\ 
ClinTox & AUROC & 0.759 & \textbf{0.818} \\ 
herg central & AUROC & 0.876 & \textbf{0.880} \\ 
hERG Karim & Accuracy & \textbf{0.728} & 0.724 \\ 
ToxCast & AUROC & 0.719 & \textbf{0.779} \\ 
SARSCoV2 Vitro Touret & AUROC & \textbf{0.550} & 0.512 \\ 
SARSCOV2 3CLPro Diamond & AUROC & \textbf{0.765} & 0.755 \\ 
HIV & AUROC & 0.668 & \textbf{0.686} \\ 
SAbDab Chen & AUPRC & \textbf{0.415} & 0.390 \\ 
HuRI & AUPRC & 0.703 & \textbf{0.705} \\ 
miRTarBase & Accuracy & \textbf{0.765} & 0.765 \\ 
MHC1 IEDB IMGT Nielsen & AUROC & 0.912 & \textbf{0.913} \\ 
MHC2 IEDB Jensen & AUROC & \textbf{0.786} & 0.781 \\ 
weber & AUROC & 0.738 & \textbf{0.738} \\ 
phase1 & AUROC & 0.608 & \textbf{0.624} \\ 
phase2 & AUROC & 0.635 & \textbf{0.639} \\ 
phase3 & AUROC & 0.691 & \textbf{0.701} \\ 
butkiewicz & AUROC & \textbf{0.621} & 0.574 \\
\bottomrule
\end{tabular}
}
\end{table}

\newcolumntype{P}[1]{>{\centering\arraybackslash}p{#1}}

\begin{table}[htbp]
\centering
\caption{Performances on regression and generation datasets with \ourmodels using 10-shot KNN prompting, with and without context. The best performances are bolded.}
\label{tab:regression_generation_nocontext}
\centerline{
\begin{tabular}{lccc}
\toprule
Dataset name & Metric & 10-shot KNN no context & 10-shot KNN with context\\
\midrule
Caco2 Wang & MAE & 0.713 & \textbf{0.621} \\ 
Lipophilicity AstraZeneca & MAE & \textbf{0.765} & 0.779 \\ 
Solubility AqSolDB & MAE & 1.101 & \textbf{0.931} \\ 
PPBR AZ & MAE & 34.763 & \textbf{11.138} \\ 
VDss Lombardo & Spearman & 0.489 & \textbf{0.497} \\ 
Half Life Obach & Spearman & 0.389 & \textbf{0.525} \\ 
Clearance Hepatocyte AZ & Spearman & 0.219 & \textbf{0.256} \\ 
Clearance Microsome AZ & Spearman & \textbf{0.401} & 0.385 \\ 
LD50 Zhu & MAE & 0.818 & \textbf{0.808} \\ 
USPTO Yields & Pearson & 0.041 & \textbf{0.042} \\ 
Buchwald Hartwig & Pearson & 0.655 & \textbf{0.682} \\ 
TAP & MAE & 5.711 & \textbf{5.075} \\ 
Leenay & Spearman & 0.036 & \textbf{0.048} \\ 
BindingDB kd & Pearson & 0.309 & \textbf{0.317} \\ 
BindingDB ic50 & Spearman & 0.318 & \textbf{0.326} \\ 
BindingDB ki & Pearson & 0.547 & \textbf{0.565} \\ 
BindingDB Patent & Pearson & 0.466 & \textbf{0.474} \\ 
DAVIS & MSE & \textbf{0.564} & 0.564 \\ 
KIBA & MSE & \textbf{0.704} & 0.709 \\ 
DisGeNET & MAE & 0.060 & \textbf{0.059} \\ 
GDSC1 & Pearson & 0.875 & \textbf{0.876} \\ 
GDSC2 & Pearson & 0.818 & \textbf{0.896} \\ 
DrugComb CSS & MAE & 21.656 & \textbf{14.740} \\ 
OncoPolyPharmacology & Pearson & 0.380 & \textbf{0.418} \\ 
Protein SAbDab & MAE & 1.433 & \textbf{1.432} \\ 
DrugComb HSA & MAE & 4.374 & \textbf{4.311} \\ 
DrugComb Loewe & MAE & 18.923 & \textbf{17.428} \\ 
DrugComb Bliss & MAE & 4.843 & \textbf{4.425} \\ 
DrugComb ZIP & MAE & 5.315 & \textbf{4.047} \\ 
USPTO & Generation Accuracy & \textbf{0.221} & 0.220 \\
\bottomrule
\end{tabular}
}
\end{table}

\begin{table}[htbp]
\centering
\caption{Performance of \ourmodels finetuned on different datasets and evaluated with 10-shot KNN prompting on binary classification datasets. "All datasets" indicates a \ourmodels model finetuned on all TDC datasets, "molecule datasets" indicates a \ourmodels model finetuned on datasets containing molecules (datasets only involving other drug types such as proteins or nucleic acids are not included in training), and "ADMET datasets" indicates a \ourmodels model finetuned on datasets in the ADMET benchmark, which only contains molecules.}
\label{tab:drug_only_binary}
\centerline{
\begin{tabular}{lP{3cm}P{1.5cm}P{1.5cm}P{1.2cm}P{1.2cm}P{1.2cm}}
\toprule
Dataset name & Feature type & In ADMET & Metric & All datasets & Molecule datasets & ADMET datasets \\
\midrule
HIA Hou & SMILES & Yes & AUROC & \textbf{0.942} & 0.928 & 0.915 \\ 
Pgp Broccatelli & SMILES & Yes & AUROC & 0.909 & \textbf{0.917} & 0.852 \\ 
Bioavailability Ma & SMILES & Yes & AUROC & 0.605 & 0.667 & \textbf{0.713} \\ 
BBB Martins & SMILES & Yes & AUROC & 0.805 & \textbf{0.860} & 0.843 \\ 
CYP2D6 Veith & SMILES & Yes & AUPRC & \textbf{0.605} & 0.522 & 0.418 \\ 
CYP3A4 Veith & SMILES & Yes & AUPRC & \textbf{0.800} & 0.737 & 0.736 \\ 
CYP2C9 Veith & SMILES & Yes & AUPRC & \textbf{0.750} & 0.698 & 0.609 \\ 
CYP2C9 Substrate CarbonMangels & SMILES & Yes & AUPRC & 0.403 & \textbf{0.455} & 0.328 \\ 
CYP2D6 Substrate CarbonMangels & SMILES & Yes & AUPRC & 0.643 & 0.670 & \textbf{0.683} \\ 
CYP3A4 Substrate CarbonMangels & SMILES & Yes & AUROC & 0.637 & \textbf{0.740} & 0.683 \\ 
hERG & SMILES & Yes & AUROC & \textbf{0.879} & 0.857 & 0.753 \\ 
AMES & SMILES & Yes & AUROC & \textbf{0.785} & 0.726 & 0.672 \\ 
DILI & SMILES & Yes & AUROC & \textbf{0.727} & 0.628 & 0.524 \\ 
phase1 & SMILES + Text & No & AUROC & \textbf{0.624} & 0.556 & 0.534 \\ 
phase2 & SMILES + Text & No & AUROC & \textbf{0.639} & 0.522 & 0.503 \\ 
phase3 & SMILES + Text & No & AUROC & \textbf{0.701} & 0.477 & 0.492 \\ 
PAMPA NCATS & SMILES & No & AUROC & 0.646 & 0.702 & \textbf{0.717} \\ 
CYP2C19 Veith & SMILES & No & AUROC & \textbf{0.877} & 0.829 & 0.797 \\ 
CYP1A2 Veith & SMILES & No & AUPRC & \textbf{0.906} & 0.861 & 0.607 \\ 
Skin Reaction & SMILES & No & AUROC & 0.564 & \textbf{0.627} & 0.388 \\ 
Carcinogens Lagunin & SMILES & No & Accuracy & 0.857 & \textbf{0.875} & 0.714 \\ 
Tox21 & SMILES & No & AUROC & \textbf{0.858} & 0.811 & 0.662 \\ 
ClinTox & SMILES & No & AUROC & \textbf{0.818} & 0.778 & 0.411 \\ 
herg central & SMILES & No & AUROC & \textbf{0.880} & 0.840 & 0.663 \\ 
hERG Karim & SMILES & No & Accuracy & \textbf{0.724} & 0.691 & 0.654 \\ 
ToxCast & SMILES & No & AUROC & \textbf{0.779} & 0.758 & 0.632 \\ 
SARSCoV2 Vitro Touret & SMILES & No & AUROC & 0.512 & \textbf{0.571} & 0.570 \\ 
SARSCOV2 3CLPro Diamond & SMILES & No & AUROC & \textbf{0.755} & 0.715 & 0.690 \\ 
HIV & SMILES & No & AUROC & \textbf{0.686} & 0.633 & 0.421 \\ 
butkiewicz & SMILES & No & AUROC & 0.574 & \textbf{0.645} & 0.498 \\ 
miRTarBase & Nucleotide + Amino acid & No & Accuracy & \textbf{0.765} & 0.499 & 0.502 \\ 
SAbDab Chen & Amino acid & No & AUPRC & 0.390 & 0.694 & \textbf{0.735} \\ 
HuRI & Amino acid & No & AUPRC & \textbf{0.705} & 0.513 & 0.513 \\ 
MHC1 IEDB IMGT Nielsen & Amino acid & No & AUROC & \textbf{0.913} & 0.675 & 0.650 \\ 
MHC2 IEDB Jensen & Amino acid & No & AUROC & \textbf{0.781} & 0.718 & 0.714 \\ 
weber & Amino acid & No & AUROC & \textbf{0.738} & 0.689 & 0.688 \\
\bottomrule
\end{tabular}
}
\end{table}

\begin{table}[htbp]
\centering
\caption{Performance of \ourmodels finetuned on different datasets and evaluated with 10-shot KNN prompting on regression and generation datasets. "All datasets" indicates a \ourmodels model finetuned on all TDC datasets, "molecule datasets" indicates a \ourmodels model finetuned on datasets containing molecules (datasets only involving other drug types such as proteins or nucleic acids are not included in training), and "ADMET datasets" indicates a \ourmodels model finetuned on datasets in the ADMET benchmark, which only contains molecules.}
\label{tab:drug_only_regression_generation}
\centerline{
\begin{tabular}{lP{3cm}P{1.5cm}P{1.5cm}P{1.2cm}P{1.2cm}P{1.2cm}}
\toprule
Dataset name & Feature type & In ADMET & Metric & All datasets & Molecule datasets & ADMET datasets \\
\midrule
Caco2 Wang & SMILES & Yes & MAE & 0.621 & 0.578 & \textbf{0.555} \\ 
Lipophilicity AstraZeneca & SMILES & Yes & MAE & \textbf{0.779} & 0.815 & 0.813 \\ 
Solubility AqSolDB & SMILES & Yes & MAE & 0.931 & 0.989 & \textbf{0.929} \\ 
PPBR AZ & SMILES & Yes & MAE & \textbf{11.138} & 11.394 & 11.441 \\ 
VDss Lombardo & SMILES & Yes & Spearman & \textbf{0.497} & 0.338 & 0.412 \\ 
Half Life Obach & SMILES & Yes & Spearman & \textbf{0.525} & 0.316 & 0.267 \\ 
Clearance Hepatocyte AZ & SMILES & Yes & Spearman & 0.256 & 0.154 & \textbf{0.262} \\ 
Clearance Microsome AZ & SMILES & Yes & Spearman & 0.385 & 0.354 & \textbf{0.519} \\ 
LD50 Zhu & SMILES & Yes & MAE & 0.808 & 0.749 & \textbf{0.713} \\ 
GDSC1 & SMILES + Text & No & Pearson & \textbf{0.876} & 0.075 & 0.146 \\ 
GDSC2 & SMILES + Text & No & Pearson & \textbf{0.896} & 0.120 & 0.135 \\ 
DrugComb CSS & SMILES + Text & No & MAE & \textbf{14.740} & 32.394 & 26.871 \\ 
OncoPolyPharmacology & SMILES + Text & No & Pearson & \textbf{0.418} & 0.097 & 0.159 \\ 
DrugComb HSA & SMILES + Text & No & MAE & \textbf{4.311} & 10.883 & 5.127 \\ 
DrugComb Loewe & SMILES + Text & No & MAE & 17.428 & 26.516 & \textbf{12.242} \\ 
DrugComb Bliss & SMILES + Text & No & MAE & \textbf{4.425} & 11.984 & 5.603 \\ 
DrugComb ZIP & SMILES + Text & No & MAE & \textbf{4.047} & 11.367 & 5.351 \\ 
USPTO Yields & SMILES & No & Pearson & \textbf{0.042} & 0.015 & 0.007 \\ 
Buchwald Hartwig & SMILES & No & Pearson & \textbf{0.682} & 0.283 & 0.554 \\ 
USPTO & SMILES & No & Generation Accuracy & \textbf{0.220} & 0.158 & 0.000 \\ 
Leenay & Nucleotide & No & Spearman & \textbf{0.048} & -0.006 & -0.014 \\ 
DisGeNET & Amino acid + Text & No & MAE & \textbf{0.059} & 0.511 & 0.432 \\ 
BindingDB kd & Amino acid + SMILES & No & Pearson & \textbf{0.317} & 0.126 & -0.001 \\ 
BindingDB ic50 & Amino acid + SMILES & No & Spearman & \textbf{0.326} & 0.004 & 0.156 \\ 
BindingDB ki & Amino acid + SMILES & No & Pearson & \textbf{0.565} & -0.057 & -0.049 \\ 
BindingDB Patent & Amino acid + SMILES & No & Pearson & \textbf{0.474} & 0.010 & -0.030 \\ 
DAVIS & Amino acid + SMILES & No & MSE & \textbf{0.564} & 14.955 & 5.631 \\ 
KIBA & Amino acid + SMILES & No & MSE & \textbf{0.709} & 5.950 & 3.391 \\ 
TAP & Amino acid & No & MAE & 5.075 & 6.491 & \textbf{5.024} \\ 
Protein SAbDab & Amino acid & No & MAE & 1.432 & 1.864 & \textbf{1.067} \\
\bottomrule
\end{tabular}
}
\end{table}

\newcolumntype{P}[1]{>{\centering\arraybackslash}p{#1}}
\begin{table}[htbp]
\centering
\caption{\color{black}The percent of each dataset's test set containing features that also exist in the PaLM-2 training data, excluding datasets with no overlap at all.}
\label{tab:percent_contamination}
\centerline{
\begin{tabular}{cP{4cm}P{2cm}P{2cm}P{2cm}}
\toprule
Dataset name & Percent overlap with PaLM-2 training data & Metric & Unfiltered performance & Filtered performance \\
\midrule
TAP & 10.42 & MAE & 4.983 & 4.560 \\
HuRI & 5.93 & AUPRC & 0.753  & 0.756 \\
SAbDab Chen & 5.39 & AUPRC & 0.473 & 0.529 \\
phase3 & 0.28 & AUROC & 0.723 & 0.727 \\
BindingDB kd & 0.09 & Pearson & 0.391 & 0.391 \\
miRTarBase & 0.04 & Accuracy & 0.804 & 0.799 \\
DisGeNET & 0.02 & MAE & 0.057 & 0.057 \\
\bottomrule
\end{tabular}
}
\end{table}

\newpage

\begin{figure}
    \centering
    \includegraphics[width=0.55\textwidth]{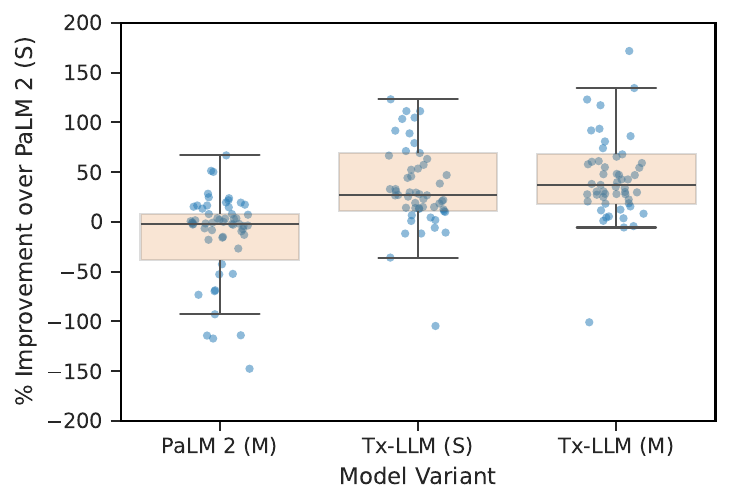}
    \caption{The percent improvement of \basemodelm, \ourmodels, and \ourmodelm compared to \basemodels for all TDC tasks. }
    \label{fig:ablations_viz_1}
\end{figure}

\begin{figure}
    \centering
    \includegraphics[width=0.6\textwidth]{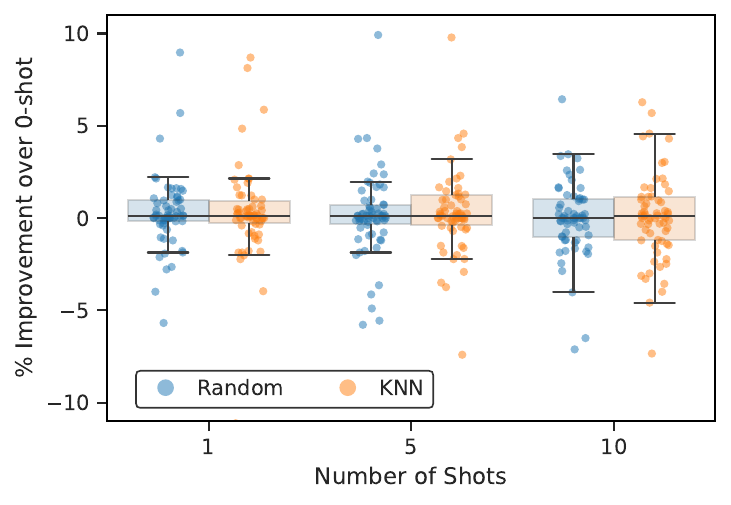}
    \caption{The percent improvement of few-shot prompting over 0-shot prompting with \ourmodels. The number of shots and the shot selection method (random or KNN) are varied.}
    \label{fig:ablations_viz_2}
\end{figure}

\newpage
\setlength\bibitemsep{3pt}
\clearpage
\printbibliography
\balance
\clearpage
\end{refsection}

\end{document}